\documentclass[journal]{IEEEtran}
\usepackage[utf8]{inputenc}
\usepackage{multicol}
\usepackage{multirow}
\usepackage[keeplastbox]{flushend}
\usepackage{diagbox}
\usepackage{amsmath}
\usepackage{booktabs}
\usepackage{multirow}
\usepackage{verbatim}
\usepackage{array}
\usepackage{dblfloatfix}
\usepackage{fixltx2e}
\usepackage{algorithmic}
\usepackage{array}
\usepackage{url}
\usepackage{float}
\usepackage{graphicx,psfrag,epsfig}
\usepackage{caption}
\usepackage{subfigure}
\usepackage{color}
\usepackage[export]{adjustbox}
\usepackage{cite}
\usepackage{amssymb}
\usepackage[ruled]{algorithm2e}

\begin{document}
\title{DGNet: Distribution Guided Efficient Learning for Oil Spill Image Segmentation} 
\author{Fang Chen, Heiko Balzter, Feixiang Zhou, Peng Ren and Huiyu Zhou
\thanks{Fang Chen, Feixiang Zhou and Huiyu Zhou are with School of Computing  
and Mathematical Sciences, University of Leicester, Leicester LE1 7RH, United Kingdom (H. Zhou is the corresponding author. E-mail:  hz143@leicester.ac.uk).}
\thanks{Heiko Balzter is with Institute for Environmental Futures, School of Geography, Geology and Environment, University of Leicester, Space Park Leicester, 92 Corporation Road, Leicester, LE4 5SP, United Kingdom.}
\thanks{Peng Ren is with College of Oceanography and Space Informatics, China University of Petroleum (East China), Qingdao 266580, China.}
}

 
\maketitle

\begin{abstract} 
Successful implementation of oil spill segmentation in Synthetic Aperture Radar (SAR) images is vital for marine environmental protection. In this paper, we develop an effective segmentation framework named DGNet, which performs oil spill segmentation by incorporating the intrinsic distribution of backscatter values in SAR images. Specifically, our proposed segmentation network is constructed with two deep neural modules running in an interactive manner, where one is the inference module to achieve latent feature variable inference from SAR images, and the other is the generative module to produce oil spill segmentation maps by drawing the latent feature variables as inputs. Thus, to yield accurate segmentation, we take into account the intrinsic distribution of backscatter values in SAR images and embed it in our segmentation model. The intrinsic distribution originates from SAR imagery, describing the physical characteristics of oil spills. In the training process, the formulated intrinsic distribution guides efficient learning of optimal latent feature variable inference for oil spill segmentation. The efficient learning enables the training of our proposed DGNet with a small amount of image data. This is economically beneficial to oil spill segmentation where the availability of oil spill SAR image data is limited in practice. Additionally, benefiting from optimal latent feature variable inference, our proposed DGNet performs accurate oil spill segmentation. We evaluate the segmentation performance of our proposed DGNet with different metrics, and experimental evaluations demonstrate its effective segmentations.
\end{abstract}

\section{introduction}
\IEEEPARstart{M}{arine} oil spills are the release of liquid petroleum hydrocarbon from oil tankers, offshore platforms or drilling rigs into the ocean environment. The occurrence of oil spills normally presents different scales and results in heavy pollution of marine waters \cite{zhang2019marine}. Oil pollution on the ocean surface may have disastrous consequences for the marine environment \cite{farrington2014oil} \cite{kingston2002long}, especially the marine ecosystem \cite{hjermann2007fish} \cite{krupp2008impact}. Thus, for marine environmental protection, it is vital to conduct timely damage assessment and spread control of oil spills \cite{soukissian2016satellite} \cite{li2017wind}. Satellite-based synthetic aperture radar (SAR) is regarded as a powerful tool for remotely observing environmental and natural targets on the earth surface because it has the all-time operation ability and independent of weather conditions \cite{velotto2016first} \cite{ulaby2014microwave}. In this case, SAR operates effectively in providing means for monitoring marine oil spills \cite{solberg2007oil}. Besides, in the operation of dealing with oil pollution, the segmentation of oil spill SAR images provides fundamental guidance and quantitative assessment for oil pollution cleanup. Thus, it is important to develop effective segmentation techniques for accurate oil spill SAR image segmentation in practice.

Researchers from geoscience and remote sensing community continue to develop strategies to analyse oil spills in SAR images, supported by their physical characteristics. Particularly, the polarimetric characteristics that contribute to oil spill outlining in SAR images have been delicately investigated, and representative studies include Salberg et al. \cite{salberg2014oil},  Migliaccio et al. \cite{buono2016polarimetric} \cite{velotto2011dual}, Espeset et al. \cite{espeseth2017analysis}, and Minchew et al. \cite{minchew2012polarimetric}. This helps image processing techniques such as graph theory \cite{gemme2018automatic} and thresholding \cite{lupidi2017fast} to better distinguish oil spills from the background. In the implementation of polarimetric SAR image segmentation, a segmentation strategy constructed with specific statistical distributions that describe SAR image data \cite{yin2014modified} may improve the segmentation performance.

On the other hand, in the image processing community, most state-of-the-art segmentation techniques have been established or explored with respect to segmentation energy functional. For example, Chen et al. \cite{chen2017level} presented an edge sensitive segmentation scheme in which the self-guided filter is incorporated in the construction of the segmentation energy functional. Xu et al. \cite{xu2017level} employed an arbitrary pixel within one marine oil spill region as initialisation to formulate the segmentation energy functional for oil spill segmentation. Marques et al. \cite{marques2011sar} exploited the splitted region information to formulate image segmentation energy functional. Jing et al. \cite{jing2011novel} defined the segmentation energy functional based on the global minimisation of an active contour model, which performs oil spill detection without local minima. Chen et al. \cite{chen2018segmenting} exploited the alternating direction method of multipliers to develop the segmentation energy functional which achieves oil spill segmentation from blurry SAR images. These strategies conduct oil spill segmentation by fitting the region information with active contour models, in which a labelled region is required for starting segmentation. Thus, the segmentation performance depends on the initial region captured \cite{ren2018energy}. However, the initial labelling is normally manually devised, which captures the initial region at a random manner. Therefore, the initial labelling is coarse, neither reliable nor efficient for capturing initial regions \cite{bertacca2005farima}, and thus the manual initialisation is hard to capture proper regions for oil spill segmentation.   

To implement effective oil spill SAR image segmentation without requiring labelled initialisation regions, in this work, we present an automatic oil spill SAR image segmentation network named DGNet, taking the advantage of the intrinsic distribution of backscatter values in SAR images. Specifically, the segmentation network composes of two deep neural modules running cooperatively in an interactive  manner, with one is the inference module that learns the representation of latent feature variables from SAR images, and the other is the generative module that generates accurate oil spill segmentation maps by feeding with the inference outputs. In practical training process, the computation for inference representation is computationally intractable. To address this issue, standard methods approximate the inference representation with a distribution that is randomly selected from a family of distributions \cite{kingma2019introduction}. The approximation with randomly selected distribution results in uncertainty for representing image inference, and thus degrades image segmentation performance. Therefore, to conduct optimal inference of SAR images to achieve effective oil spill segmentation. Different from the standard techniques, in our proposed DGNet, we incorporate the intrinsic distribution of backscatter values in SAR images into our segmentation model. The intrinsic distribution originates from SAR imagery for maritime scenes, providing physical characteristics for describing oil spills. Thus, in the training for segmentation, the incorporated intrinsic distribution guides efficient learning of optimal latent feature variable inference for oil spill segmentation. The efficient learning enables that successful training of our proposed DGNet without requiring large number of oil spill SAR image data. This is economically beneficial to oil spill segmentation where the availability of oil spill SAR image data is limited in practice. Furthermore, the optimal latent feature variable inference enables that our proposed DGNet performs effective segmentation for oil spill areas with irregular shapes. The main contributions of our oil spill SAR image segmentation approach are summarised as follows:

$\bullet$ We propose a novel oil spill SAR image segmentation scheme by addressing latent feature variable inference in SAR images. We achieve this objective using the intrinsic distribution of backscatter values in SAR images, which provides physical characteristics of oil spills and thus guides efficient learning of optimal latent feature variable inference.

$\bullet$ We construct the segmentation network with two deep neural modules that work cooperatively in an interactive manner, and the training of these two modules are simultaneously undertaken. Benefiting from the efficient learning, the training of our proposed DGNet does not require a large amount of image data. This leads to the less demand of training samples for oil spill SAR image segmentation where the availability of SAR image data is actually limited in many cases.

$\bullet$ We conduct SAR image segmentation with optimal latent feature variable inference. This enables our proposed DGNet to accurately delineate irregularly shaped oil spill areas. Comprehensive evaluation validates the proposed model. 


Extensive experimental evaluations are conducted to validate the performance of our proposed segmentation framework, and the evaluations with different metrics validate its effectiveness over different of oil spill scenes.

\section{Related Work}
\label{neural net representation for image segmentation}
We construct our proposed oil spill SAR image segmentation network by taking the recent advances of deep neural networks with probabilistic analysis in an explicit way. Evidence has shown that explicit probabilistic representations help to efficiently deliver image segmentation tasks \cite{pu2016variational}. In oil spill SAR images, oil spill areas appear as dark patches, and thus correct characterising dark patches is critical for oil spill SAR image segmentation \cite{shu2010dark}. Thus, to represent oil spill SAR image segmentation, in this section, we briefly review the process of conducting image inference for segmentation, whereas the inference representation is approximated by a distribution randomly selected from a family of distributions.

Specifically, to explicitly deal with image segmentation problems, two modules are established to support each other, one for latent variable inference and the other for producing segmentation maps. Thus, to depict SAR image segmentation. Let $\mathcal{I}$ be the input SAR image data, $\Omega \rightarrow \mathbb{R}$ be the image domain, and $\mathcal{S}^{LV}$ be the inferred latent variables. According to \cite{kingma2019introduction} \cite{fox2012tutorial}, we derive the following inference representation: 
\begin{equation}
  P_{ind}(\mathcal{S}^{LV}\mid \mathcal{I})=\frac{P_{jnt}(\mathcal{I}, \mathcal{S}^{LV})}{P_{pri}(\mathcal {I})}
  \label{inference distribution}
\end{equation}
where $P_{jnt}( \mathcal{I}, \mathcal{S}^{LV})$ is the joint distribution representation of latent variable $\mathcal{S}^{LV}$ and image data $\mathcal{I}$,  $P_{pri}(\mathcal{I})$ is the prior distribution of the image data, and $P_{ind}(\mathcal{S}^{LV}\!\!\mid\! \mathcal{I})$ is the distribution representation of the inference outcomes conditioned on image data $\mathcal{I}$. Particularly, in the segmentation network, the inference and segmentation are internal correlated and thus to describe the interaction of inference and segmentation, we firstly write the joint distribution as follows:
\begin{equation}
P_{jnt}(\mathcal{I}, \mathcal{S}^{LV})=P_{ged}(\mathcal {I}\mid \mathcal{S}^{LV})P_{pri}(\mathcal{S}^{LV})
\label{intermediate transfer}
\end{equation}
where $P_{ged}(\mathcal {I}\!\mid \!\mathcal{S}^{LV})$ and $P_{pri}(\mathcal{S}^{LV}\!)$ represent the distributions of the generative process conditioned on $\mathcal{S}^{LV}$ and the prior of $\mathcal{S}^{LV}\!\!$, respectively. From Eq. (\ref{intermediate transfer}), we rewrite Eq. (\ref{inference distribution}) that links with the generative process as follows:
\begin{equation}
  P_{ind}(\mathcal{S}^{LV}\mid \mathcal{I})=\frac{P_{ged}(\mathcal {I}\mid \mathcal{S}^{LV})P_{pri}(\mathcal{S}^{LV})}{P_{pri}(\mathcal{I})} 
  \label{inference representation}
\end{equation}
In the segmentation module, the segmentation is operated by drawing the latent variables from the inference outcomes, and thus the segmentation performance heavily relies on the inference of the latent variables. To specifically represent the inference and the segmentation operations. Examining Eq. (\ref{inference representation}), we calculate the denominator $P_{pri}(\mathcal{I})$ by marginalising out $\mathcal{S}^{LV}$ as follows:
\begin{equation}
  P_{pri}(\mathcal{I})=\int_{\Omega}P_{ged}(\mathcal {I}\mid \mathcal{S}^{LV})P_{pri}(\mathcal{S}^{LV})d\mathcal{S}^{LV}
  \label{computation of input image}
\end{equation}
where the integration computation is achieved over all the configurations of $\mathcal{S}^{LV}$, and thus it requires exponential time to deliver this computation, which is computationally intractable and may degrade the segmentation in practice. To address this concern, a distribution $Q_{\tau}^{seld}(\mathcal{S}^{LV}\!\!\!\mid \!\mathcal{I})$, which is randomly selected from a family of distributions \cite{fox2012tutorial}, is introduced to approximate the inference distribution $ P_{ind}(\mathcal{S}^{LV}\!\!\mid \mathcal{I})$. To examine the similarity between these two distributions, the Kullback-Leibler (KL) divergence \cite{hershey2007approximating} is employed, which measures the information loss when using $Q_{\tau}^{seld}$ to approximate $P_{ind}$, and the measurement is given as follows:
\begin{equation}
\begin{split}
&D_{\mathbb{KL}}(Q_{\tau}^{seld}(\mathcal{S}^{LV}\mid \mathcal{I})\parallel P_{ind}(\mathcal{S}^{LV}\mid \mathcal{I}))=\\
&\int_{\Omega} Q_{\tau}^{seld}(\mathcal{S}^{LV}\mid \mathcal{I}){\rm log}\frac{Q_{\tau}^{seld}(\mathcal{S}^{LV}\mid \mathcal{I})}{P_{ind}(\mathcal{S}^{LV}\mid \mathcal{I})}d\mathcal{S}^{LV} 
\end{split}
\label{first kl divergence}
\end{equation}
Since the distribution is randomly selected from a family of distributions, and thus to make these two distributions approach, the minimisation computation for Eq. (\ref{first kl divergence}) is conducted, which is shown as follows:
\begin{equation}
\begin{split}
{Q^{*}}_{\tau}^{seld}(\mathcal{S}^{LV}\mid \mathcal{I})=&{\rm arg\ min}D_{\mathbb{KL}}(Q_{\tau}^{seld}(\mathcal{S}^{LV}\\
&\mid\mathcal{I})\parallel P_{ind}(\mathcal{S}^{LV}\mid \mathcal{I}))
\end{split}
\label{optimization for the approximation distribution}
\end{equation}

In terms of the intractable computation arising in the inference process, a randomly selected distribution is used to approximate the representation of the inference outputs. This brings much uncertainty in representing image inferences. The uncertainty of image inference representation further results in inaccurate image segmentation. On the other hand, in maritime scenes, due to turbulence and wind blowing on the ocean surface, oil spills in SAR images normally exhibit various areas with irregular shapes. This bringings challenge for accurate segmentation, and there is no established model to exactly outline oil spills in SAR images. To address this issue and render effective oil spill segmentation, according to the work of inference learning \cite{cremer2018inference}  \cite{kim2018semi} and latent characterisation \cite{mathieu2019disentangling} \cite{tang2019correlated}, and with the exploration of the distribution representation for oil spill SAR images, we here present a novel segmentation technique for oil spill SAR image segmentation.

\begin{figure*}[h]
\begin{center}
\includegraphics[width=1.08\textwidth,height=0.433\textheight,center]{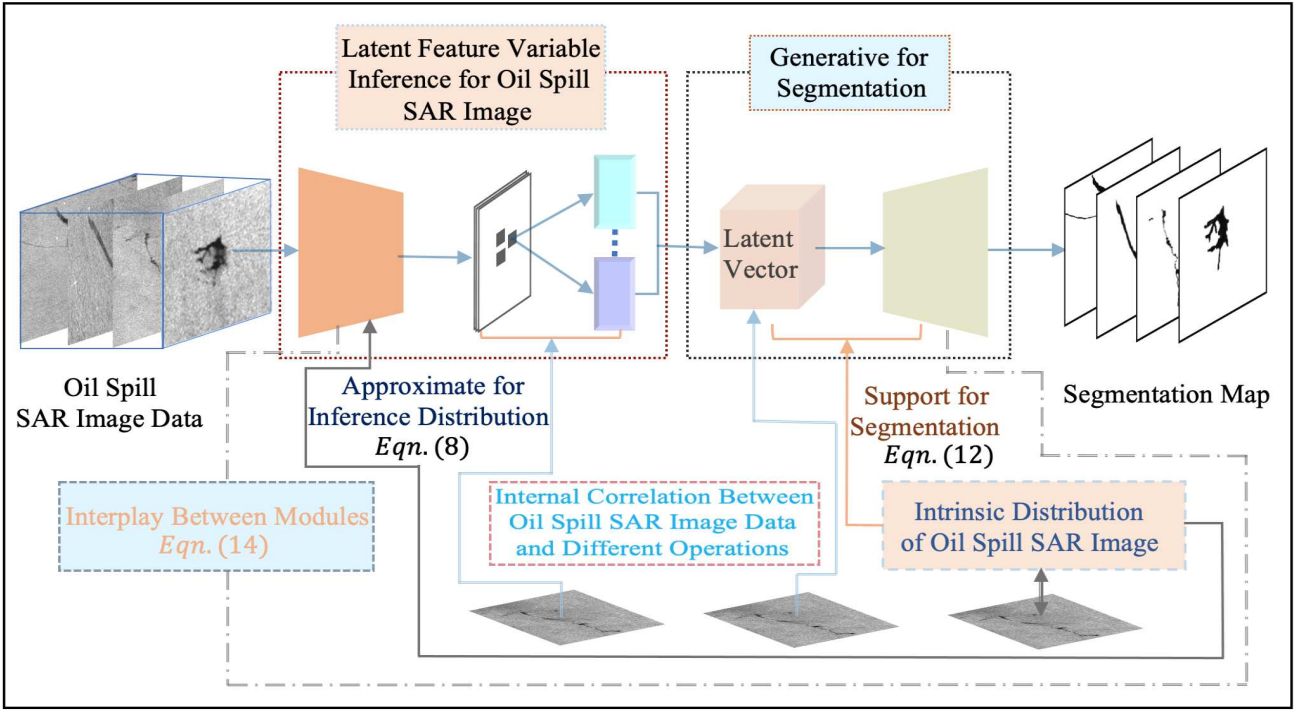}
\end{center}
\vspace*{-4.7mm}
\caption{The architecture of the proposed oil spill SAR image segmentation framework. The proposed segmentation framework is constructed with an inference module and a generative module that work to conduct oil spill SAR image segmentation in an interplay manner. Specifically, the inference module is structured to operate inference for the input oil spill SAR images, and the generative module is structured to generate oil spill segmentation maps by drawing the latent feature variables from the inference outputs. Thus, to render effective oil spill segmentation, we incorporate the intrinsic distribution of backscatter values in SAR images into our segmentation model. The intrinsic distribution originates from SAR imagery for maritime scenes, providing physical characteristics of oil spills and thus guiding efficient learning for effective oil spill segmentation.}
\label{the oil spill segmentation framework}
\end{figure*}

\begin{figure}[htbp]
\begin{center}
\includegraphics[width=0.33\textwidth,height=0.20\textheight,center]{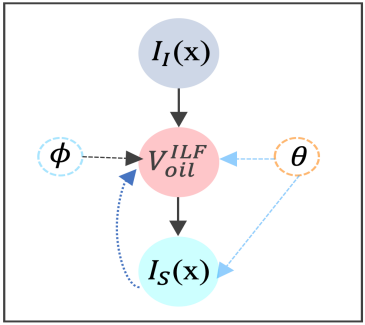}
\end{center}
\vspace*{-4.7mm}
\caption{Graph model illustration of the segmentation scheme. In this graph, $I_{I}(\textbf x)$, $V_{oil}^{ILF}$ and $I_{S}(\textbf x)$ represent the input image, latent feature variable and the segmentation map, respectively. $\phi$ and $\theta$ indicate the distributions in the segmentation network, which are jointly learned in the training process.}
\label{the graph model framework}
\end{figure}
  %

\section{Construction of Distribution Guided Efficient Learning for Oil Spill SAR Image Segmentation}
\label{the construction of oil spill SAR image segmentation network}
To render effective oil spill segmentation, in this section, we present the processes of constructing an effective segmentation method to perform accurate oil spill segmentation. This segmentation method incorporates SAR image intrinsic distribution to guide optimal latent feature variable inference for effective oil spill segmentation. Particularly, in probability theory and statistics, the intrinsic distribution refers to a mathematical function that describes the statistical representation of datapoints, where the physical characteristics for describing data formation are modelled \cite{hung2014intrinsic}. In the field of machine learning, a latent feature is such feature that is not directly observed but can be extracted through a mathematical model, and it is used to predict the results \cite{nguyen2015improving}. Thus, the intrinsic distribution and latent feature are two vital components in the construction of effective image segmentation techniques. Thus, to accurately segment oil spills in SAR images, different from standard methods which approximate the inference representation with a randomly selected distribution, we here construct our segmentation model with the intrinsic distribution of backscatter values in SAR images embedded. The intrinsic distribution originates from SAR imagery for maritime scenes, which provides physical characteristics of oil spills. Therefore, the incorporated intrinsic distribution guides efficient learning of optimal latent feature variable inference, enabling effective oil spill segmentation without requiring a large amount of oil spill SAR image data to train the segmentation model. This provides an economical way for oil spill segmentation where the availability of oil spill SAR image data is scare in practice. The proposed segmentation framework is illustrated in Fig. \ref{the oil spill segmentation framework}.

\subsection{Intrinsic Distribution of Backscatter values in Oil Spill SAR Image}   
\label{oil spill probability distribution representation}
To construct our segmentation framework by taking into account the physical characteristics of oil spills, we here investigate the intrinsic distribution of backscatter values in oil spill SAR images. Particularly, let us denote $\mathcal{I}$ as an oil spill SAR image, and $\mathcal{I}(\textbf x)$ as the image vector. Based on the description of SAR imagery for maritime scenes and the established models for SAR image representation \cite{ulaby2014microwave} \cite{oliver2004understanding} \cite{gao2019characterization}, we produce the intrinsic distribution of backscatter values in oil spill SAR images below:
\begin{equation}
\mathcal{P}_{itd}(\mathcal{I}(\textbf x))=\frac{1}{K_{s}\sigma(\textbf x)}\rm exp^{(-\frac{1}{K_{s}\sigma(\textbf x)}\mathcal{I}(\textbf x))}, \mathcal{I}(\textbf x)\geq0
\label{oil spill distribution functional}
\end{equation}
where $K_{s}$ is the detection system constant, and $\sigma$ is the radar cross section component which is information bearing of SAR image pixels, and thus the information in each pixel is carried by $\sigma$ \cite{oliver2004understanding}. Therefore, the intrinsic distribution represents the statistics of image pixels, presenting a more detailed characterisation for SAR images. The segmentation work conducted in \cite{chen2021oil} reported that a proper distribution utilised in the construction of the segmentation model benefits to effectively segment oil spills from the background in SAR images. Thus, to render effective segmentation, we here incorporate the derived intrinsic distribution into our segmentation network for guiding optimal latent feature variable inference for implementing accurate oil spill segmentation.

\subsection{Efficient Learning via Intrinsic Distribution of Backscatter Values in SAR Images}
\label{efficient learning for oil spill segmentation}
In the implementation of deep learning based image segmentation, one assumption is to request a large number of images for training the segmentation networks, but this pre-requisite normally cannot be satisfied in oil spill segmentation tasks since the availability of oil spill SAR image data is scare in practice. To address this and render accurate oil spill segmentation, in this work, we integrate the intrinsic distribution shown in Eq. (\ref{oil spill distribution functional}) into our segmentation model. The integrated intrinsic distribution is expected to facilitate efficient learning for optimal latent feature variable inference for performing accurate segmentations.

Specifically, our proposed DGNet consists of an inference module and a generative module, which act cooperatively to conduct the segmentation work. In the training process, the inference module learns the representation of latent feature variable inference, and the generative module produces oil spill segmentation maps using latent feature variable as inputs. The graph model of our proposed DGNet is illustrated in Fig. \ref{the graph model framework}. Thus, to conduct optimal latent feature variable inference for accurate oil spill segmentation, taking advantage of the intrinsic distribution that represents SAR images shown in Section\ref{oil spill probability distribution representation}, we here describe how to incorporate the intrinsic distribution into our segmentation model for SAR image segmentation. Particularly, let $\mathcal{I}(\textbf x)$ denote the input SAR image data, and $\mathcal{V}_{oil}^{ILF}$ denote the vector of the latent feature variables derived with the intrinsic distribution embedded. In the training process, we have the distribution representation of the inference outputs shown as $\mathcal{D}_{\theta}^{ifd}(\mathcal{V}_{oil}^{ILF}\!\!\mid \!\!\mathcal{I}(\textbf x))$, and the intrinsic distribution with respect to oil spill SAR image characterisation is $\mathcal{P}_{\phi}^{itd}(\mathcal{V}_{oil}^{ILF}\!\!\!\mid\! \mathcal{I}(\textbf x))$. In dealing with intractable computation arises in the inference representation, standard methods introduce a distribution that is randomly selected from a family of distributions to approximate the distribution of the inference outputs. The approximation with randomly selected distribution results in uncertainty for oil spill SAR image inference, and thus degenerates the segmentation performance. To address this and render accurate oil spill segmentation, we incorporate the intrinsic distribution $\mathcal{P}_{\phi}^{itd}(\mathcal{V}_{oil}^{ILF}\!\!\!\mid\! \mathcal{I}(\textbf x))$ into our segmentation model. To measure the closeness between the inference and the intrinsic distributions, KL divergence is exploited, given below:
\begin{equation}
\begin{split}
 &D_{\mathbb {KL}}\big[\mathcal{P}_{\phi}^{itd}\big(\mathcal{V}_{oil}^{ILF}\mid \mathcal{I}(\textbf x)\big)\parallel \mathcal{D}_{\theta}^{ifd}\big(\mathcal{V}_{oil}^{ILF}\mid \mathcal{I}(\textbf x)\big)\big]=\\
 &\int_{\Omega}\mathcal{P}_{\phi}^{itd}\big(\mathcal{V}_{oil}^{ILF}\mid \mathcal{I}(\textbf x)\big){\rm log}\frac{\mathcal{P}_{\phi}^{itd}\big(\mathcal{V}_{oil}^{ILF}\mid \mathcal{I}(\textbf x)\big)}{\mathcal{D}_{\theta}^{ifd}\big(\mathcal{V}_{oil}^{ILF}\mid \mathcal{I}(\textbf x)\big)}d\mathcal{V}_{oil}^{ILF}
 \end{split}
 \label{kl divergence for approximate and true distributions}
\end{equation}
where $\Omega \subset \Re ^{2}$ is the SAR image domain, $\phi$ and $\theta$ are the parameters related to the segmentation network. As described in our segmentation network, the inference and the generative modules work cooperatively in an interactive manner, and the segmentation performance depends on the latent feature variable inference. Thus, to measure the divergence between the inference and the intrinsic distributions for segmentation, we rewrite the log denominator $\mathcal{D}_{\theta}^{ifd}\big(\mathcal{V}_{oil}^{ILF}\mid \mathcal{I}(\textbf x)\big)$ that connects with the generative process on the right hand side (RHS) of Eq. (\ref{kl divergence for approximate and true distributions}) as follows:
\begin{equation}
\begin{split}
  \mathcal{D}_{\theta}^{ifd}\big(\mathcal{V}_{oil}^{ILF}\mid\mathcal{I}(\textbf x)\big)=\frac{\mathcal{D}_{\theta}^{ged}\big(\mathcal{I}(\textbf x)\mid \mathcal{V}_{oil}^{ILF} \big)\mathcal{D}_{\theta}^{pri}\big(\mathcal{V}_{oil}^{ILF}\big)}{\mathcal{D}_{\theta}^{pri}\big(\mathcal{I}(\textbf x)\big)}
  \end{split}
  \label{bayesian transforation}
\end{equation}
where $\mathcal{D}_{\theta}^{ged}(\mathcal{I}(\textbf x)\!\!\mid\!\! \mathcal{V}_{oil}^{ILF})$ indicates the distribution representation of the generative process, and this distribution is conditioned on the inferred latent feature variable $\mathcal{V}_{oil}^{ILF}$. Thus, Eq. (\ref{bayesian transforation}) establishes the correlation between the inference and the generative modules in the segmentation network. In combination with Eq. (\ref{bayesian transforation}), we derive the computation for the RHS of Eq. (\ref{kl divergence for approximate and true distributions}) as follows: 
\begin{equation}
\begin{split}
&\!\!\!\!\int_{\Omega}\!\!\! \mathcal{P}_{\phi}^{itd}\big(\mathcal{V}_{oil}^{ILF}\!\!\mid\! \mathcal{I}(\textbf x)\big){\rm log}\frac{\mathcal{P}_{\phi}^{itd}\big(\mathcal{V}_{oil}^{ILF}\!\!\mid\! \mathcal{I}(\textbf x)\big)}{\mathcal{D}_{\theta}^{ifd}\big(\mathcal{V}_{oil}^{ILF}\!\!\mid\! \mathcal{I}(\textbf x)\big)}d\mathcal{V}_{oil}^{ILF}\!\!=\!\!\!\int_{\Omega}\!\!\!\! \mathcal{P}_{\phi}^{itd}\\
&\!\!\!\!\big(\mathcal{V}_{oil}^{ILF}\!\!\mid\! \mathcal{I}(\textbf x)\!\big){\rm log}\frac{\mathcal{P}_{\phi}^{itd}\big(\mathcal{V}_{oil}^{ILF}\!\!\mid\! \mathcal{I}(\textbf x)\big)\mathcal{D}_{\theta}^{pri}\big(\mathcal{I}(\textbf x)\big)}{\mathcal{D}_{\theta}^{ged}\big(\mathcal{I}(\textbf x)\!\!\mid \!\mathcal{V}_{oil}^{ILF} \big)\mathcal{D}_{\theta}^{pri}\big(\mathcal{V}_{oil}^{ILF}\big)}d\mathcal{V}_{oil}^{ILF}
\end{split}
\label{specific calculation for the divergence}
\end{equation}
This equation brings together the logarithm and the integration computations. To conduct detailed computations for this equation, we commence by calculating the integration with respect to $\mathcal{P}_{\phi}^{itd}(\mathcal{V}_{oil}^{ILF}\!\mid\! \mathcal{I}(\textbf x))$ and the log numerator on the RHS of Eq. (\ref{specific calculation for the divergence}) as follows:
\begin{equation}
\begin{split}
&\int_{\Omega}\!\! \mathcal{P}_{\phi}^{itd}\big(\mathcal{V}_{oil}^{ILF}\!\mid\! \mathcal{I}(\textbf x)\!\big)
{\rm log}\mathcal{P}_{\phi}^{itd}\big(\mathcal{V}_{oil}^{ILF}\!\mid\! \mathcal{I}(\textbf x)\!\big)
\mathcal{D}_{\theta}^{pri}\big(\mathcal{I}(\textbf x)\!\big)\\
&\!\!d\mathcal{V}_{oil}^{ILF}\!\!\!=\!\!\!\int_{\Omega}\!\!\!\mathcal{P}_{\phi}^{itd}\big(\mathcal{V}_{oil}^{ILF}\!\mid\! \mathcal{I}(\textbf x)\!\big){\rm log}\mathcal{P}_{\phi}^{itd}\big(\mathcal{V}_{oil}^{ILF}\!\mid\! \mathcal{I}(\textbf x)\!\big)d\mathcal{V}_{oil}^{ILF}\\
&\!\!+\!\!\int_{\Omega}\!\!\mathcal{P}_{\phi}^{itd}\big(\mathcal{V}_{oil}^{ILF}\!\mid\! \mathcal{I}(\textbf x)\big){\rm log}\mathcal{D}_{\theta}^{pri}\big(\mathcal{I}(\textbf x)\big)d\mathcal{V}_{oil}^{ILF}\!\!=\!{\rm log}\mathcal{D}_{\theta}^{pri}\\
&\!\!\big(\mathcal{I}(\textbf x)\big)\!\!+\!\!\! \int_{\Omega}\!\!\!\mathcal{P}_{\phi}^{itd}\big(\mathcal{V}_{oil}^{ILF}\!\!\mid\!\! \mathcal{I}(\textbf x)\big)
{\rm log}\mathcal{P}_{\phi}^{itd}\big(\mathcal{V}_{oil}^{ILF}\!\!\mid\! \mathcal{I}(\textbf x)\big)d\mathcal{V}_{oil}^{ILF}
\end{split}
\label{numerator calculation}
\end{equation}
This formulation shows the detailed computation for the integration of the intrinsic distribution and the log numerator in Eq. (\ref{specific calculation for the divergence}). Examining the RHS of Eq. (\ref{specific calculation for the divergence}), the log computation is a fraction, and thus we need to compute the denominator component. The integration computation that corresponds to the intrinsic distribution  $\mathcal{P}_{\phi}^{itd}(\mathcal{V}_{oil}^{ILF}\!\mid\! \mathcal{I}(\textbf x))$ and the log denominator on the RHS of Eq. (\ref{specific calculation for the divergence}) is shown as follows:
\begin{equation}
\begin{split}
&\!\!\int_{\Omega}\!\! \mathcal{P}_{\phi}^{itd}\big(\mathcal{V}_{oil}^{ILF}\!\mid\! \mathcal{I}(\textbf x)\!\big){\rm log}\mathcal{D}_{\theta}^{ged}\big(\mathcal{I}(\textbf x)\!\mid\! \mathcal{V}_{oil}^{ILF} \!\big)
\mathcal{D}_{\theta}^{pri}\big(\mathcal{V}_{oil}^{ILF}\big)d\mathcal{V}_{oil}^{ILF}\\
&\!=\!\!\int_{\Omega}\!\! \mathcal{P}_{\phi}^{itd}\big(\mathcal{V}_{oil}^{ILF}\!\mid\! \mathcal{I}(\textbf x)\!\big){\rm log}\mathcal{D}_{\theta}^{ged}\big(\mathcal{I}(\textbf x)\mid \mathcal{V}_{oil}^{ILF} \big)d\mathcal{V}_{oil}^{ILF}\!\!+\!\!\int_{\Omega}\mathcal{P}_{\phi}^{itd}\\
&\big(\mathcal{V}_{oil}^{ILF}\!\mid\! \mathcal{I}(\textbf x)\big)
{\rm log} \mathcal{D}_{\theta}^{pri}\big(\mathcal{V}_{oil}^{ILF}\big)d\mathcal{V}_{oil}^{ILF}=\!\!\int_{\Omega}\!\!\mathcal{P}_{\phi}^{itd}\big(\mathcal{V}_{oil}^{ILF}\!\mid\! \mathcal{I}(\textbf x)\big)\\
&\quad {\rm log} \mathcal{D}_{\theta}^{pri}\big(\mathcal{V}_{oil}^{ILF}\big)d\mathcal{V}_{oil}^{ILF}+\mathbb{E}_{\mathcal{P}_{\phi}^{itd}}\big[{\rm log}\mathcal{D}_{\theta}^{ged}\big(\mathcal{I}(\textbf x)\mid \mathcal{V}_{oil}^{ILF}\big)\big]
\end{split}
\label{denominator calculation}
\end{equation}
the last term in the above equation is the expectation with respect to the generative process, which is undertaken over the representation of the intrinsic distribution. In our segmentation network, the incorporated intrinsic distribution guides efficient learning for optimal latent feature variable inference whilst supporting effective segmentation. Eqs. (\ref{numerator calculation}) and (\ref{denominator calculation}) represent the computations for Eq. (\ref{specific calculation for the divergence}). For clarity, we combine the integration components of these two equations as follows: 
\begin{equation}
\begin{split}
&\int_{\Omega}\mathcal{P}_{\phi}^{itd}\big(\mathcal{V}_{oil}^{ILF}\mid \mathcal{I}(\textbf x)\big)
{\rm log}\mathcal{P}_{\phi}^{itd}\big(\mathcal{V}_{oil}^{ILF}\mid \mathcal{I}(\textbf x)\big)d\mathcal{V}_{oil}^{ILF}\\
&\quad -\!\!\int_{\Omega}\mathcal{P}_{\phi}^{itd}\big(\mathcal{V}_{oil}^{ILF}\mid \mathcal{I}(\textbf x)\big){\rm log} \mathcal{D}_{\theta}^{pri}\big(\mathcal{V}_{oil}^{ILF}\big)d\mathcal{V}_{oil}^{ILF}\\
&\qquad=D_{\mathbb{KL}}\big[\mathcal{P}_{\phi}^{itd}\big(\mathcal{V}_{oil}^{ILF}\mid \mathcal{I}(\textbf x)\big)\parallel \mathcal{D}_{\theta}^{pri}\big(\mathcal{V}_{oil}^{ILF}\big)\big]
\end{split}
\end{equation}
Using the above equation, we have the detailed computation for the similarity measurement between the intrinsic distribution and the inference distribution in Eq. (\ref{kl divergence for approximate and true distributions}), shown as follows:
\begin{equation}
\begin{split}
&D_{\mathbb {KL}}\big[\mathcal{P}_{\phi}^{itd}\big(\mathcal{V}_{oil}^{ILF}\mid \mathcal{I}(\textbf x)\big)\parallel\mathcal{D}_{\theta}^{ifd}(\mathcal{V}_{oil}^{ILF}\mid \mathcal{I}(\textbf x))\big]\\
&=\!\int_{\Omega}\!\mathcal{P}_{\phi}^{itd}\big(\mathcal{V}_{oil}^{ILF}\mid \mathcal{I}(\textbf x)\big){\rm log}\mathcal{P}_{\phi}^{itd}\big(\mathcal{V}_{oil}^{ILF}\mid \mathcal{I}(\textbf x)\big)d\mathcal{V}_{oil}^{ILF}\\
&-\!\!\int_{\Omega}\!\mathcal{P}_{\phi}^{itd}\big(\mathcal{V}_{oil}^{ILF}\mid \mathcal{I}(\textbf x)\big){\rm log}\mathcal{D}_{\theta}^{ifd}\big(\mathcal{V}_{oil}^{ILF}\mid\! \mathcal{I}(\textbf x)\big)d\mathcal{V}_{oil}^{ILF}\!=\!\\
&{\rm log}\mathcal{D}_{\theta}^{pri}\big(\mathcal{I}(\textbf x)\!\big)\!\!+\!\! D_{\mathbb{KL}}\!\big[\mathcal{P}_{\phi}^{itd}\big(\mathcal{V}_{oil}^{ILF}\!\mid\! \mathcal{I}(\textbf x)\big)\!\parallel\! \mathcal{D}_{\theta}^{pri}\big(\mathcal{V}_{oil}^{ILF}\big)\big]\\
&\quad \qquad -\mathbb{E}_{\mathcal{P}_{\phi}^{itd}}\big[{\rm log}\mathcal{D}_{\theta}^{ged}\big(\mathcal{I}(\textbf x)\mid \mathcal{V}_{oil}^{ILF}\big)\big]
\end{split}
\label{final representation of the kl divergence}
\end{equation}
The above equation correlates the computations for both the inference and the generative processes, and within them the intrinsic distribution is involved. Thus, in our proposed segmentation network, the embedded intrinsic distribution contributes to the latent feature variable inference and the generative segmentation stage. As shown above that the log distribution representation of the generative process in the expectation is conditioned on the inferred latent feature variables. This describes the interactive between the inference and the generative modules in our segmentation network, and the segmentation performance is closely correlated to the outcomes of the inference outputs. Therefore, in the training for oil spill SAR image segmentation, to enable accurate segmentation, it is critical to conduct the optimised computation for the segmentation network, which is described in the next section. 

\begin{algorithm}[htbp]
\caption{The Procedures of Implementing Marine Oil Spill SAR Image Segmentation.}
\label{training for segmenting oil spill framework}
\KwIn{Marine oil spill image dataset including the original oil spill images and the ground-truth segmentations.}
\KwOut{Oil spill segmentation outcomes.}
\begin{algorithmic}[1] 
    \STATE Initialise parameters $\phi$ and $\theta$;
    \STATE \textbf {while} in iteration numbers \textbf {do}
    \STATE Draw mini-batch examples from the training dataset;
    \STATE Optimise the lower bound $\mathcal{L}(\phi, \theta)$ w.r.t  both $\phi$ and $\theta$;
    \STATE $\quad$Train to minimise Eq. (\ref{KL divergence of inference}) encouraging the \\
    $\quad$  distributions $\mathcal{P}_{\phi}^{itd}\big(\mathcal{V}_{oil}^{ILF}\mid \mathcal{I}(\textbf x^{(i)})\big)$ and $\mathcal{D}_{\theta}^{pri}\big(\mathcal{V}_{oil}^{ILF}\big)$ \\
    $\quad$ to move closely;
    \STATE $\quad$ Draw samples from the prior $\mathcal{D}_{\theta}^{pri}\big(\mathcal{V}_{oil}^{ILF}\big)$;
    \STATE $\quad$ Train to maximise Eq. (\ref{expectation for encoder net});
    \STATE Update parameters $\phi$ and $\theta$ using gradient descent method;
    \STATE \textbf {end while};
\RETURN Trained parameters $\phi$, $\theta$;
\end{algorithmic}
\end{algorithm}

\section{Optimisation for Effective Oil Spill SAR Image Segmentation}
We present the details of incorporating the intrinsic distribution of backscatter values in SAR images to construct the segmentation network in Section \ref{the construction of oil spill SAR image segmentation network}. Specifically, our proposed segmentation network is structured with an inference module and a generative module running in an interactive manner, where the inference module learns the representation of optimal latent feature variable inference for SAR images that characterise oil spill areas precisely, and the generative module takes the inferred latent feature variables as inputs to generate accurate oil spill segmentation maps in the training procedure. Thus, to conduct optimal latent feature variable inference for effective segmentation, we look at the difference measurement between the inference distribution and the intrinsic distribution, and when the inference distribution tightly approaches the intrinsic distribution, the inference module achieves optimal latent feature variable inference, and this benefits to conduct accurate oil spill segmentations. 

To effectively handle the above problem, we optimise the KL divergence minimisation by maximising its corresponding evidence lower bound (ELBO). ELBO is used to describe the KL divergence minimisation in a computational tractable style. Thus, to demonstrate the capability of ELBO for KL divergence minimisation, we also investigate other alternative methods, according to \cite{hu2013kullback}, we understand that expectation and chance constrained methods are feasible options as well. Particularly, the expectation constrained method exploits the parameters in the objective function to formulate an expectation constraint term, while the chance constrained method designs the model constraint term with a constraint function satisfying a pre-defined probability. Therefore, these methods result in additional computational complexity and thus influence the efficiency of optimisation. Therefore, in this paper, we use ELBO instead of expectation/chance constrained methods because of computational efficiency.

Particularly, for an input oil spill SAR image $\mathcal{I}(\textbf x)$ with $N$ datapoints in the image domain $\Omega$ (i.e. $\mathcal{I}(\textbf x)=\{\mathcal{I}(\textbf x^{(i)})\}_{i=1}^{N}$), we represent the difference measurement between the inference and the intrinsic distributions in Eq. (\ref{final representation of the kl divergence}) with respect to individual datapoint as follows:
\begin{equation}
\begin{split}
&D_{\mathbb {KL}}\big[\mathcal{P}_{\phi}^{itd}\big(\mathcal{V}_{oil}^{ILF}\mid \mathcal{I}(\textbf x^{(i)})\big)\parallel \mathcal{D}_{\theta}^{ifd}\big(\mathcal{V}_{oil}^{ILF}\mid \mathcal{I}(\textbf x^{(i)})\big)\big]\\
&={\rm log}\mathcal{D}_{\theta}^{pri}\big(\mathcal{I}(\textbf x^{(i)})\big)\!-\mathbb{E}_{\mathcal{P}_{\phi}^{itd}}\big[{\rm log}\mathcal{D}_{\theta}^{ged}\big(\mathcal{I}(\textbf x^{(i)})\!\mid\!\mathcal{V}_{oil}^{ILF}\big)\big]\\
&\quad +D_{\mathbb{KL}}\big[\mathcal{P}_{\phi}^{itd}\big(\mathcal{V}_{oil}^{ILF}\mid \mathcal{I}(\textbf x^{(i)})\big)\parallel \mathcal{D}_{\theta}^{pri}\big(\mathcal{V}_{oil}^{ILF}\big)\big]
\end{split}
\label{tractable calculation for kl divergence}
\end{equation}
The above equation describes the similarity measurement between the intrinsic distribution $\mathcal{P}_{\phi}^{itd}(\mathcal{V}_{oil}^{ILF}\!\!\mid\!\!\mathcal{I}(\textbf x^{(i)}\!)\!)$ and the inference distribution $\mathcal{D}_{\theta}^{ifd}(\!\mathcal{V}_{oil}^{ILF}\!\!\mid\!\!\mathcal{I}(\textbf x^{(i)}\!)\!)$ from each datapoint, which ensures more accurate latent feature variable inference for effective segmentation. Therefore, in the training for oil spill SAR image segmentation, we perform the optimisation computation by Eq. (\ref{tractable calculation for kl divergence}). Specifically, for optimisation, examining the component $\mathcal{D}_{\theta}^{pri}\!(\mathcal{I}(\textbf x^{(i)})\!)$ shown in Eq. (\ref{tractable calculation for kl divergence}), we marginalise out the latent feature variable $\mathcal{V}_{oil}^{ILF}$ as follows:
\begin{equation}
\mathcal{D}_{\theta}^{pri}\big(\mathcal{I}(\textbf x^{(i)})\big)\!=\!\!\int_{\Omega}\!\mathcal{D}_{\theta}^{ged}\big(\mathcal{I}(\textbf x^{(i)})\mid \mathcal{V}_{oil}^{ILF}\big)\mathcal{D}_{\theta}^{pri}\big(\mathcal{V}_{oil}^{ILF}\big)d\mathcal{V}_{oil}^{ILF}
\label{integration computation}
\end{equation}
According to the aforementioned description that the integration computation requires exponential time, and this results in the computation for $\mathcal{D}_{\theta}^{pri}(\mathcal{I}(\textbf x^{(i)}))$ is intractable. In this scenario, a straightforward optimisation computation for Eq. (\ref{tractable calculation for kl divergence}) is difficult in practice. To address this and render tractable optimisation computation, we rewrite Eq. (\ref{tractable calculation for kl divergence}) that regards to ${\rm log}\mathcal{D}_{\theta}^{pri}(\mathcal{I}(\textbf x^{(i)}))$ as follows:
\begin{equation}
\begin{split}
&{\rm log}\mathcal{D}_{\theta}^{pri}\big(\mathcal{I}(\textbf x^{(i)})\big)=D_{\mathbb {KL}}\big[\mathcal{P}_{\phi}^{itd}\big(\mathcal{V}_{oil}^{ILF}\mid \mathcal{I}(\textbf x^{(i)})\big)\parallel\\
&\mathcal{D}_{\theta}^{ifd}\big(\mathcal{V}_{oil}^{ILF}\mid \mathcal{I}(\textbf x^{(i)})\big)\big]-D_{\mathbb{KL}}\big[\mathcal{P}_{\phi}^{itd}\big(\mathcal{V}_{oil}^{ILF}\mid\mathcal{I}(\textbf x^{(i)})\big)\\
&\parallel\mathcal{D}_{\theta}^{pri}\big(\mathcal{V}_{oil}^{ILF}\big)\big]+\mathbb{E}_{\mathcal{P}_{\phi}^{itd}}\big[{\rm log}\mathcal{D}_{\theta}^{ged}\big(\mathcal{I}(\textbf x^{(i)})\mid \mathcal{V}_{oil}^{ILF}\big)\big]
\end{split}
\label{variational evidence lower bound}
\end{equation}
This represents that ${\rm log}\mathcal{D}_{\theta}^{pri}(\mathcal{I}(\textbf x^{(i)}))$ in Eq. (\ref{tractable calculation for kl divergence}) can be solved in a different way, and the computation for ${\rm log}\mathcal{D}_{\theta}^{pri}(\mathcal{I}(\textbf x^{(i)}))$ is equivalent to derive the components on the RHS of Eq. (\ref{variational evidence lower bound}). To avoid cumbersome representation, we further represent the last two terms on the RHS of Eq. (\ref{variational evidence lower bound}) as follows:
\begin{equation}
\begin{split}
 &\mathcal{L}(\phi, \theta) =\mathbb{E}_{\mathcal{P}_{\phi}^{itd}}\big[{\rm log}\mathcal{D}_{\theta}^{ged}\big(\mathcal{I}(\textbf x^{(i)})\mid \mathcal{V}_{oil}^{ILF}\big)\big]\\
 &-D_{\mathbb{KL}}\big[\mathcal{P}_{\phi}^{itd}\big(\mathcal{V}_{oil}^{ILF}\mid \mathcal{I}(\textbf x^{(i)})\big)\parallel\mathcal{D}_{\theta}^{pri}\big(\mathcal{V}_{oil}^{ILF}\big)\big]
\end{split}
\label{segmentation evidence lower bound}
\end{equation}
Therefore, the representation for ${\rm log}\mathcal{D}_{\theta}^{pri}(\mathcal{I}(\textbf x^{(i)}))$ in Eq. (\ref{variational evidence lower bound}) is equivalent to the following equation:
\begin{equation}
\begin{split}
&{\rm log}\mathcal{D}_{\theta}^{pri}\big(\mathcal{I}(\textbf x^{(i)})\big)=\mathcal{L}(\phi, \theta)+D_{\mathbb {KL}}\big[\mathcal{P}_{\phi}^{itd}\\
&\big(\mathcal{V}_{oil}^{ILF}\mid \mathcal{I}(\textbf x^{(i)})\big)
\parallel\mathcal{D}_{\theta}^{ifd}\big(\mathcal{V}_{oil}^{ILF}\mid \mathcal{I}(\textbf x^{(i)})\big)\big]
\end{split}
\label{inequality with respect to the evidence lower bound}
\end{equation}
By Jensen's inequality \cite{chandler1988introduction}\cite{kuczma2009introduction}, the  KL divergence is non-negative. Thus, we have the correlation between ${\rm log}\mathcal{D}_{\theta}^{pri}(\mathcal{I}(\textbf x^{(i)}))$ and  $\mathcal{L}(\phi, \theta)$ shown as follows:
\begin{equation}
{\rm log}\mathcal{D}_{\theta}^{pri}\big(\mathcal{I}(\textbf x^{(i)})\big)\geq\mathcal{L}(\phi, \theta)
\label{representation of the evidence lower bound}
\end{equation}
which demonstrates that $\mathcal{L}(\phi, \theta)$ is the lower bound of ${\rm log}\mathcal{D}_{\theta}^{pri}\big(\mathcal{I}(\textbf x^{(i)})\big)$, and thus $\mathcal{L}(\phi, \theta)$ is called the ELBO. From Eqs. (\ref{inequality with respect to the evidence lower bound}) and (\ref{representation of the evidence lower bound}), we have the minimisation for $D_{\mathbb {KL}}\big[\mathcal{P}_{\phi}^{itd}\big(\mathcal{V}_{oil}^{ILF}\!\!\mid \!\!\mathcal{I}(\textbf x^{(i)})\big)\!\!\parallel\!\!\mathcal{D}_{\theta}^{ifd}\big(\mathcal{V}_{oil}^{ILF}\!\!\mid\!\! \mathcal{I}\!(\textbf x^{(i)}\!)\!\big)\!\big]$ as maximising the ELBO $\mathcal{L}(\phi, \theta)$, which is computationally tractable. Therefore, we reformulate the KL divergence minimisation problem into the ELBO maximisation problem as follows:
\begin{equation}
\begin{split}
{\rm min} D_{\mathbb {KL}}&\big[\mathcal{P}_{\phi}^{itd}\big(\mathcal{V}_{oil}^{ILF}\!\mid\! \mathcal{I}(\textbf x^{(i)})\big)\!\parallel\!\mathcal{D}_{\theta}^{ifd}
\big(\mathcal{V}_{oil}^{ILF}\mid\mathcal{I}(\textbf x^{(i)})\big)\big]\\
&\simeq {\rm max}\mathcal{L}(\phi, \theta)
\end{split}
\end{equation}
Taking a close look at $\mathcal{L}(\phi, \theta)$ in Eq. (\ref{segmentation evidence lower bound}), the maximisation for $\mathcal{L}(\phi, \theta)$ is achieved by maximising the first term whilst the second term is minimised, and thus the $\mathcal{L}(\phi, \theta)$ maximisation is undertaken in two steps: one is the min-step as follows:
\begin{equation}
\min_{\phi}  D_{\mathbb{KL}}\big[\mathcal{P}_{\phi}^{itd}\big(\mathcal{V}_{oil}^{ILF}\mid \mathcal{I}(\textbf x^{(i)})\big)\parallel\mathcal{D}_{\theta}^{pri}\big(\mathcal{V}_{oil}^{ILF}\big)\big]
\label{KL divergence of inference}
\end{equation}
The other is the max-step as follows:
\begin{equation}
\max_{\theta} \mathbb{E}_{\mathcal{P}_{\phi}^{itd}}\big[{\rm log}\mathcal{D}_{\theta}^{ged}\big(\mathcal{I}(\textbf x^{(i)})\mid \mathcal{V}_{oil}^{ILF}\big)\big]
\label{expectation for encoder net}
\end{equation}
Thus, in the training for oil spill SAR image segmentation, the latent feature variable inference for generating accurate segmentation maps is conducted by maximising the ELBO estimator $\mathcal{L}(\phi, \theta)$, and the maximisation for $\mathcal{L}(\phi, \theta)$ is achieved with the min-max computation in Eqs. (\ref{KL divergence of inference}) and (\ref{expectation for encoder net}). 

The illustration of our proposed oil spill SAR image segmentation approach is described in Algorithm \ref{training for segmenting oil spill framework}.

\section{Experimental Work}
In this section, we conduct experimental evaluations for our proposed marine oil spill SAR image segmentation framework by comparing its segmentation against several state-of-the-art segmentation methodologies. Specifically, to validate the performance of our proposed method for oil spill SAR image segmentation, especially the segmentation for irregular oil spill areas, we exploit different types of oil spill SAR images as our experimental dataset, and the segmentation evaluations from different metrics are shown in the following parts.
\subsection{Experimental Preparations}
\label{description for image dataset}
\textit{\textbf{Image Dataset}}: In the experimental work, we exploit the oil spill SAR images obtained from the NOWPAP database\footnote{http://cearac.poi.dvo.ru/en/db/} with VV, VH and HH polarization, as our experimental dataset. Specifically, the oil spill SAR images we used in the experimental work include C-band oil spill SAR images from ERS-1 and ERS-2 satellites, C-band oil spill ASAR images from Envisat satellite, and X-band oil spill X-SAR images from X-SAR satellite. These images were captured in different time by different sensors and contain irregular oil spill areas, and we provide the specifications of the oil spill SAR images in terms of image sources and sensor properties in Tables \ref{information of SAR sensors} and \ref{SAR image information of NOWPAP}, where the symbol "-" in Table \ref{SAR image information of NOWPAP} indicates unavailable information.

\begin{table}[htbp]
	\renewcommand\arraystretch{1.96}
	\centering
	\tabcolsep 0.11in
	\caption{SATELLITE SENSORS AND IMAGE DESCRIPTION.}
	\begin{tabular}{c|c|c|c}
		\hline
		\hline
		Satellite Sensor& Spatial Resolution & Waveband & Image Level  \\
		\hline
		ERS-1 SAR & 30m x 30m  & C-band & 2\\  
		\hline
		ERS-2 SAR & 30m x 30m  & C-band & 2\\ 
		\hline	
		X-SAR & 6m x 6m  & X-band & 2 \\
		\hline
		Envisat ASAR & 150m x 150m  & C-band & 2 \\
		\hline
		\hline
	\end{tabular}
	\label{information of SAR sensors}	
\end{table}

\begin{table}[H]
	\renewcommand\arraystretch{1.97}
	\centering
	\tabcolsep 0.0023in
	\caption{DESCRIPTION OF OIL SPILL IMAGES.}
	\begin{tabular}{c|c|c|c}
		\hline
		\hline
		Capture Time & Satellite Sensor & Type of Oil Spills & Image Cover Ground \\
		\hline
		19.06.1995 02:30:40 & ERS-1 & - & 394 km$^{2}$ \\
		\hline
		19.06.1995 02:30:26 & ERS-1 & - & - \\
		\hline
		19.06.1995 02:30:12 & ERS-1 &  - & 394 km$^{2}$ \\
		\hline
		01.10.1994 05:30:54 & X-SAR & - & 9.4x106m$^{2}$ \\
		\hline
		20.07.1997 02:14:26 & ERS-2 & Ship Oil Spill & - \\ 
		\hline
		19.06.1995 02:30:12 & ERS-2 & Ship Oil Spill &- \\
		\hline
		02.09.1996 02:00:55 & ERS-2 & - &17.8x106m$^{2}$ \\
		\hline
		16.08.2007 01:16:02  & Envisat &  Ship Oil Spill &- \\
		\hline
		30.08.2006 13:04:30 & Envisat & - &-  \\
		\hline
		28.07.2008 12:26:30  & Envisat & - & -\\
		\hline
		\hline
	\end{tabular}
	\label{SAR image information of NOWPAP}	
\end{table}


\subsection{Experimental Setup}
To describe the effectiveness of our proposed segmentation network for oil spill segmentation, we conduct experiments on different types of SAR image data and the image data information is given in Section \ref{description for image dataset}. The detailed descriptions for implementing experimental work are given as follows:
\subsubsection{Implementation Details} Our proposed segmentation network performs oil spill image segmentation with the size of the input image data as 256 $\times$ 256. Specifically, in our proposed segmentation network, the module of image inference is structured with 4 convolutional layers and 1 fully-connected layer where each convolutional layer is followed by a batch normalization layer, and we use the LeakyReLU as the activation function. The module of generating segmentation maps is structured nearly an inverse form of the image inference module, where the convolutional layer is replaced with deconvolutional layer. In the segmentation network, the inference outputs are placed in the latent feature vector which is represented with distribution representation that is characterised with mean and standard deviation, and thus the number channel of the latent feature vector is 2. Particularly, the proposed segmentation network implements SAR image segmentation in an end-to-end manner, where the inference module is trained to conduct the inference for input SAR images, and the generative module is trained to generate oil spill segmentation maps by feeding with the inferred outputs, and this constructs an interactive manner for the segmentation work. 
\subsubsection{Training Operation} We perform the experimental work using the Tensorflow framework. Particularly, the segmentation model is trained on a NVIDIA Tesla P100 GPU with 16GB memory for 160 epochs with the batch size set as 1, and the training is conducted using the Adam optimizer with the learning rate set to $\alpha=1e-4$.

\begin{figure*}[t]
	\begin{center}
	\includegraphics[width=0.98\textwidth,height=0.41\textheight,center]{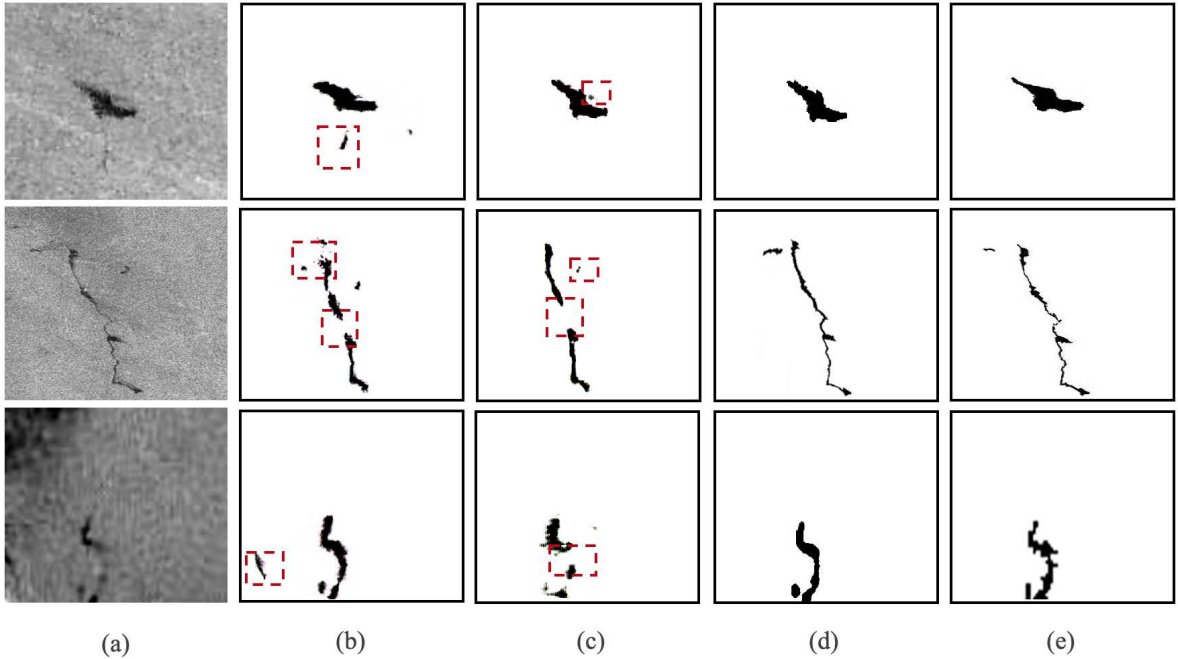}
	\end{center}
	\vspace*{-8.7mm}
	\caption{The segmentation of oil spill SAR images with the exploitation of different segmentation techniques. Specifically, (a) shows the original oil spill images, and from top to bottom are the oil spill SAR, ASAR and X-SAR images, respectively. (b)-(d) are the segmentation results of GAN, TVDL and our proposed method, respectively. (e) shows the corresponding ground truth segmentations. The red dashed-line boxes are utilised to indicate the incorrect segmentations.}
  \label{deep net segmentation comparison}
\end{figure*}

\subsection{Segmentation for Different Types of Oil Spill SAR Images}
\label{initial segmentation results comparison} 
To validate the segmentation performance of our proposed method for oil spill SAR image segmentation, in this subsection, we conduct the experimental work by comparing its segmentation with that of different state-of-the-art representative segmentation methodologies. Specifically, in this experimental work, we commence the experimental validation by comparing the segmentation results of our proposed DGNet with those of the generative adversarial network (GAN) \cite{goodfellow2014generative} and the total variation divergence learning (TVDL) strategy \cite{yu2018oil} for different types of oil spill SAR image segmentation. These two comparison techniques work in different mechanisms in terms of training for image segmentation, and taking them as comparison enables a more wide range examination on our proposed DGNet. Particularly, in the implementation of oil spill SAR image segmentation, GAN performs the segmentation by training two deep neural models that play against each other to achieve segmentation. Different from GAN, the TVDL technique performs oil spill SAR image segmentation by training the segmentation neural network in terms of minimising the disagreement between the model generated segmentations and the ground truth segmentations, whereas the disagreement between them is formulated with respect to the total variation divergence. Thus, to validate the segmentation performance of our proposed method, GAN and the TVDL techniques, we implement them to segment oil spill SAR, ASAR and X-SAR images simultaneously, and the segmentation results are shown in Fig. \ref{deep net segmentation comparison}. The segmentation results show that our proposed method achieves more accurate oil spill image segmentation compared against GAN and the TVDL segmentation techniques. This benefits from that we construct our segmentation network by incorporating the intrinsic distribution that represents oil spill SAR images into the segmentation model to guide efficient learning for optimal latent feature variable inference, and feeding the generative module with the optimal inferred latent feature variables contributes to perform accurate oil spill segmentations. Besides, to further evaluate the segmentation performance of our proposed method, we compare its segmentation with GAN and the TVDL techniques with respect to segmentation accuracy. Specifically, in image segmentation, the segmentation accuracy represents the percentage of the correctly segmented pixels \cite{cantorna2019oil}, and it is computed to represent the performance of the proposed segmentation method. The computation of the segmentation accuracy is shown below, and the results are shown in Table \ref{segmentation accuracy of deep neural techniques}. This Table shows that our proposed segmentation network achieves higher segmentation accuracy compared against GAN and the TVDL methods. The evaluations from both qualitative and quantitative metrics validate the effectiveness of our proposed segmentation method for oil spill SAR, ASAR and X-SAR image segmentation. 

\begin{equation*}
\label{accuracy computation}
\text{Accuracy} = \frac{  \text{\# the number of correctly segmented pixels}}{ \text{\# the number of all pixels}}
\end{equation*}

\begin{table}[htbp]
	\renewcommand\arraystretch{2.03}
	\begin{center}
	\tabcolsep 0.123in
	\caption{THE SEGMENTATION ACCURACY OF GAN, TVDL AND OUR PROPOSED METHOD.}
	\begin{tabular}{c|ccc}
		\hline
		\hline
		\multirow{2}{*}{}{\diagbox[innerwidth=2.3cm]{Image}{Method}}
		&GAN & TVDL &Our Method\\
		\hline
		\hline
		SAR & 0.8997 & 0.8979 & 0.9623 \\
		\hline
	    ASAR& 0.7659 & 0.8321 & 0.9867 \\
		\hline
		X-SAR & 0.8668 & 0.7506 & 0.9369 \\
		\hline
		\hline
	\end{tabular}
	\label{segmentation accuracy of deep neural techniques}
	\end{center}
	\vspace*{-4.7mm}
\end{table}

\begin{figure*}[t]
	\begin{center}
	\includegraphics[width=0.98\textwidth,height=0.42\textheight,center]{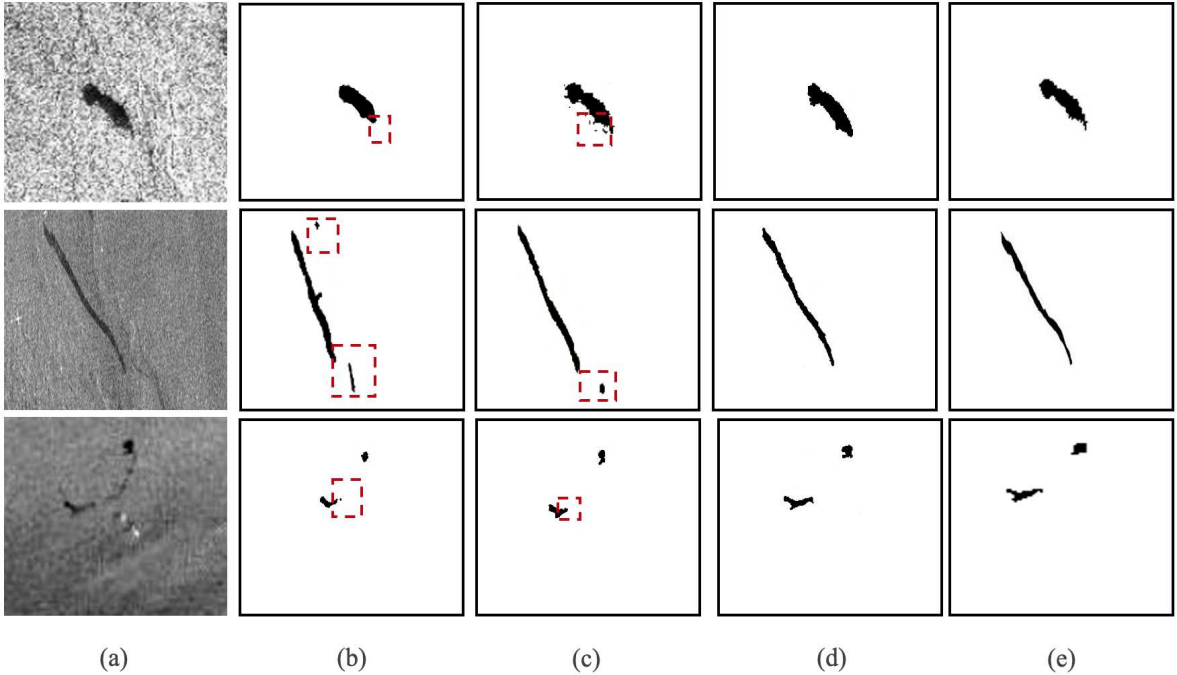}
	\end{center}
	\vspace*{-6.7mm}
	\caption{Oil spill SAR image segmentation with the exploitation of the edge-based, region-based and the proposed segmentation methods. Specifically, (a) shows the original oil spill images, and from top to bottom are the SAR, ASAR and X-SAR images, respectively. (b)-(d) are the segmentations of DRLSE, RSF and our proposed method, separately. (e) shows the corresponding ground truth segmentations. The red dashed-line boxes are utilised to indicate the incorrect segmentations.}
  \label{classical method segmentation comparison}
\end{figure*}

In addition, in the implementation of image segmentation, the segmentation techniques formulated using textual and edge information of images are considered effective for segmentation \cite{wu2017backscattering} \cite{luo2018multi}. Thus, to evaluate the segmentation performance of our proposed method one step further, we compare its segmentation with those of the region-scalable fitting (RSF) \cite{li2008minimization} and the distance regularized level set evolution (DRLSE) \cite{li2010distance} segmentation methods. Specifically, RSF is a representative region-based image segmentation method, which deals with image segmentation with intensity inhomogeneity, in which the segmentation energy functional is formulated by fitting the local region information outside and inside the segmentation contour. Different from the RSF segmentation strategy, the DRLSE is a popular edge-based segmentation method, which implements image segmentation using edge information. 

Specifically, to validate the segmentation performance of our proposed method, RSF and the DRLSE for different types of oil spill SAR image segmentation, we utilise them to segment the oil spill SAR, ASAR and X-SAR images simultaneously in the experimental work, and the segmentation results are shown in Fig. \ref{classical method segmentation comparison}. From the segmentation results, it is clear that our proposed method outperforms the RSF and the DRLSE for different experimental types of oil spill SAR image segmentation. Particularly, the RSF and the DRLSE segmentation strategies perform oil spill SAR image segmentation relying on region and edge information, respectively. However, due to turbulence and wind blowing on the ocean surface, the oil spills in SAR images normally exhibit various areas with irregular shapes. Thus, in the implementation of oil spill SAR image segmentation with the exploitation of region-based and edge-based segmentation methodologies, it is hard to devise the segmentation formulation that properly represents the region or edge information of oil spills in SAR images, and this degenerates the segmentation performance. Additionally, to quantitatively evaluate the segmentation performance of our proposed method, we compute the segmentation accuracy for our proposed method, RSF and the DRLSE for different types of oil spill SAR image segmentation, and the results are shown in Table \ref{segmentation accuracy of region-based and edge-based methods}. From this Table, it is clear that our proposed method achieves higher segmentation accuracy compared against the RSF and the DRLSE segmentation methods. The evaluations from both qualitative and quantitative metrics validate the effectiveness of our proposed method for different types of oil spill SAR image segmentation.

\begin{table}[h]
 	\renewcommand\arraystretch{2.03}
 	\begin{center}
 	\tabcolsep 0.123in
 	\caption{THE SEGMENTATION ACCURACY OF DRLSE, RSF AND OUR PROPOSED METHOD.}
 	\begin{tabular}{c|ccc}
		\hline
 		\hline
 		\multirow{2}{*}{}{\diagbox[innerwidth=2.3cm]{Image}{Method}}
		&DRLSE & RSF &Our Method\\
		\hline
 		\hline
 		SAR & 0.8769 & 0.9085 & 0.9875 \\
 		\hline
 	    ASAR& 0.8558 & 0.9129 & 0.9901 \\
 		\hline
 		X-SAR & 0.7549 & 0.7821 & 0.9667 \\
 		\hline
 		\hline
 	\end{tabular}
 	\label{segmentation accuracy of region-based and edge-based methods}
 	\end{center}
 \end{table}

The above evaluations demonstrate the effectiveness of our proposed method. Moreover, in the operation of satellite image segmentation, the F1-score and the Intersection over Union (IoU) are two useful metrics in evaluating the segmentation performance \cite{muruganandham2016semantic}. Thus, to make further evaluation on our proposed method for the segmentation of SAR, ASAR and X-SAR images, respectively, we here conduct the evaluation in terms of the F1-score and IoU, and they are defined as follows: 
\begin{equation*}
\text{F1 score} = \frac{  \text{2 $\times$ recall$\times$ precision}}{ \text{recall + precision}}
\end{equation*}
\begin{equation*}
\text{IoU} = \frac{ \text{TP}}{ \text{TP + FP + FN}}
\end{equation*}
where TP, FP and FN represent the correctly segmented oil pixels, the background pixels that are incorrectly segmented as oil, and the oil pixels that are incorrectly segmented as background, respectively. In other words, the F1-score is the harmonic mean of recall and precision, while the IoU, as its name suggests, is the intersection of the actual oil spill areas and the segmented areas, divided by the union of these two sections. The evaluations for SAR, ASAR and X-SAR image segmentation with respect to these two metrics are shown in Tables \ref{F1 score and IoU for SAR image segmentation}-\ref{F1 score and IoU for X-SAR image segmentation}. These Tables show that our proposed segmentation network achieves better results of IoU and F1 score, compared against the DRLSE and the RSF segmentation methods, and the results of these two metrics are close. Specifically,  \cite{eelbode2020optimization} shows that there is a certain relationship between IoU and F1 score, i.e. if a method performs better than the others in terms of IoU, it will also perform better in terms of F1 score, and vice-versa. Examining the experimental results shown in Tables \ref{F1 score and IoU for SAR image segmentation}-\ref{F1 score and IoU for X-SAR image segmentation}, we can obtain that the employed methods also follow such trend. For instance, in Table \ref{F1 score and IoU for SAR image segmentation}, it shows that our proposed segmentation network performs better in terms of IoU, and it also performs better in terms of F1 score compared against the DRLSE and the RSF methodologies for SAR image segmentation. Besides, benefiting from the guidance of the incorporated intrinsic distribution for effective learning, our proposed method demonstrates stable performance against varied metrics. 

In addition, in mathematical formulation, IoU and F1 score can be equivalently represented as: ${\rm IoU}=\frac{|s\cap \tilde{s}|}{|s\cup \tilde{s}|}$ and ${\rm F1}=\frac{2|s\cap \tilde{s}|}{|s|+|\tilde{s}|}$, where $|s|$ and $|\tilde{s}|$ are the generated segmentation and the ground truth segmentation, respectively. The formulation of IoU and F1 score shows that the numerator part of F1 score double the numerator of IoU, while for the denominator part, F1 score is computed with $|s|+|\tilde{s}|$, and IoU is computed with $|s\cup \tilde{s}|$, the union of these two segmentations. Thus, when the generated segmentation tightly follows the ground truth segmentation, $|s|+|\tilde{s}|$ is almost double $|s\cup \tilde{s}|$ mathematically.  This explains why our proposed method results in much closer IoU and F1 score, compared against DRLSE and RSF, shown in Tables \ref{F1 score and IoU for SAR image segmentation}-\ref{F1 score and IoU for X-SAR image segmentation}. 

\begin{table}[h]
 	\renewcommand\arraystretch{2.03}
 	\begin{center}
 	\tabcolsep 0.147in
 	\caption{THE EVALUATION OF IOU AND F1 SCORE FOR SAR IMAGE SEGMENTATION.}
 	\begin{tabular}{c|ccc}
		\hline
 		\hline
 		\multirow{2}{*}{}{\diagbox[innerwidth=2.47cm]{Evaluation metric}{Method}}
		& DRLSE & RSF & Our Method\\
		\hline
 		\hline
 		IOU & 0.8745 & 0.9073 & 0.9805 \\
 		\hline
 	    F1 SCORE & 0.8823 & 0.9104 & 0.9897 \\
 		\hline
 		\hline
 	\end{tabular}
 	\label{F1 score and IoU for SAR image segmentation}
 	\end{center}
 	\vspace*{-1.7mm}
 \end{table}
 
 \begin{table}[h]
 \renewcommand\arraystretch{2.03}
 	\begin{center}
 	\tabcolsep 0.147in
 	\caption{THE EVALUATION OF IOU AND F1 SCORE FOR ASAR IMAGE SEGMENTATION.}
 	\begin{tabular}{c|ccc}
		\hline
 		\hline
 		\multirow{2}{*}{}{\diagbox[innerwidth=2.47cm]{Evaluation metric}{Method}}
		& DRLSE & RSF & Our Method\\
		\hline
 		\hline
 		IOU & 0.8537 & 0.9074 & 0.9897 \\
 		\hline
 	    F1 SCORE & 0.8629 & 0.9203 & 0.9947 \\
 		\hline
 		\hline
 	\end{tabular}
 	\label{F1 score and IoU for ASAR image segmentation}
 	\end{center}
 	\vspace*{-1.7mm}
 \end{table}
 
 \begin{table}[h]
 	\renewcommand\arraystretch{2.03}
 	\begin{center}
 	\tabcolsep 0.147in
 	\caption{THE EVALUATION OF IOU AND F1 SCORE FOR X-SAR IMAGE SEGMENTATION.}
 	\begin{tabular}{c|ccc}
		\hline
 		\hline
 		\multirow{2}{*}{}{\diagbox[innerwidth=2.47cm]{Evaluation metric}{Method}}
		& DRLSE & RSF & Our Method\\
		\hline
 		\hline
 		IOU & 0.7544 & 0.7817 & 0.9654 \\
 		\hline
 	    F1 SCORE & 0.7953 & 0.8016 & 0.9701 \\
 		\hline
 		\hline
 	\end{tabular}
 	\label{F1 score and IoU for X-SAR image segmentation}
 	\end{center}
 	\vspace*{-1.7mm}
 \end{table}

\subsection{Comparison with Initialisation Dependent Methods}
\label{comparison with the initialisation dependent method}
\begin{figure*}[ht]
  \centering
  \subfigure[]{
  \includegraphics[width=0.97\textwidth,height=0.28\textheight]{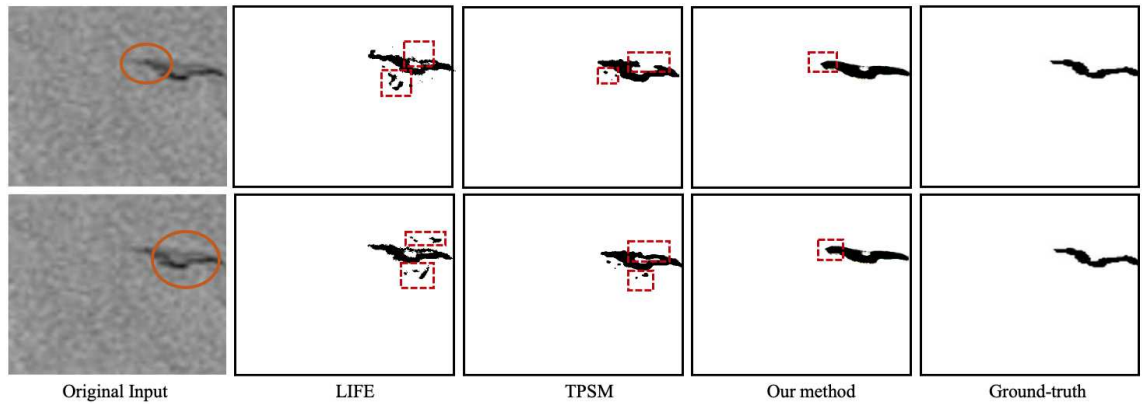}}
  \subfigure[]{
  \includegraphics[width=0.97\textwidth,height=0.28\textheight]{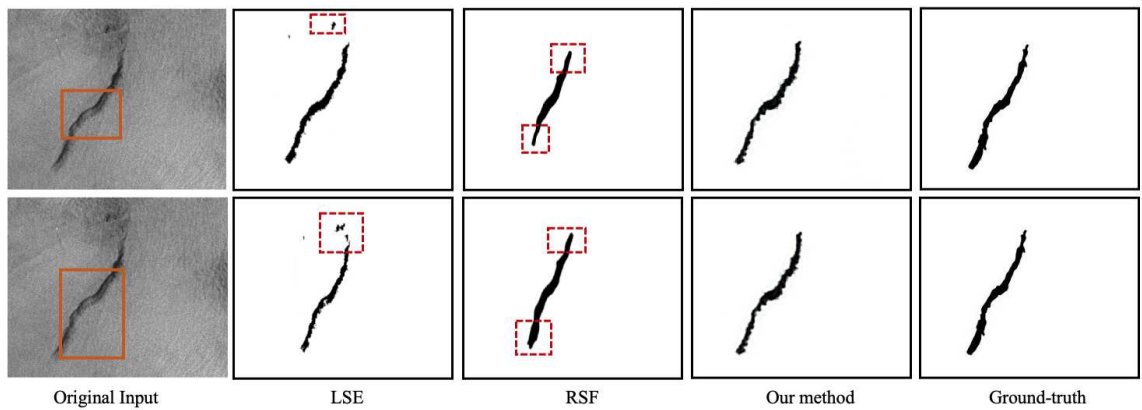}}
\caption{Comparison against the initialisation dependent methods. In this figure, (a) and (d) show the comparison against the circular initialisation dependent methods and the rectangle initialisation dependent strategies, respectively. Specifically, (a) from left to right are the input oil spill SAR image initialised with different circulars, the segmentations with LIFE, TPSM and our proposed method, and the ground-truth segmentation, respectively. (b) from left to right are the input oil spill SAR image initialised with different rectangles, the segmentations with LSE, RSF and our proposed method, and the ground-truth segmentation, respectively. The red dashed line boxes are exploited to indicate the incorrect segmentations.}
\label{comparison with initialisation dependent method}
\end{figure*}

We present an automatic marine oil spill SAR image segmentation framework, which performs the training for oil spill SAR image segmentation in an end-to-end form. Thus, to evaluate the segmentation performance of our proposed segmentation method one step further, we here compare its segmentation with the manual initialisation dependent methods. Specifically, the initialisation dependent methods operate the segmentation by manually devising an initialisation contour to capture an initialisation region for starting segmentation, and thus the segmentation performance are correlated to the devised initialisation region. Particularly, for the initialisation dependent methods, there are mainly two categories of initialisation, i.e. initialisation with a circular or initialisation with a rectangle. Thus, to operate a detailed validation for our proposed method, we compare its segmentation with representative circular initialisation methods including local image fitting energy (LIFE) \cite{zhang2010active} and two-phase segmentation model (TPSM) \cite{zhang2015level}, and the rectangle initialisation methods include RSF and the level set evolution (LSE) \cite{li2011level}. For the comparison with the circular initialisation dependent methods, to depict the segmentation with respect to the initialisation, we manually set up the initialisation with two different sized circulars at different locations, and the initialisations and the corresponding segmentation results are shown in Fig. \ref{comparison with initialisation dependent method}(a). Examining the segmentation results, it is obvious that our proposed segmentation method achieves more accurate oil spill SAR image segmentation compared against the LIFE and the TPSM circular initialisation dependent segmentation methods.  

In addition, in the operation of comparing our proposed method with rectangle initialisation dependent methods, we manually set up two different initialisation rectangles at different locations to capture the starting region, and the initialisations and the corresponding segmentation results are shown in Fig. \ref{comparison with initialisation dependent method}(b). This figure shows that for the RSF and the LSE segmentation methods, the segmentation results vary with different initialisations. The segmentation results show that our proposed method outperforms the RSF and the LSE segmentation methods for oil spill SAR image segmentation. Besides, to further demonstrate the segmentation performance of our proposed method, confusion matrix that shows the segmentation of oil spill images is given in Table \ref{confusion matrix for oil spill image segmentation}. In this Table, the diagonal cells represent the percentage of correct segmentation of oil spills and background, respectively, while the non-diagonal cells in top and bottom rows represent the percentage of oil spills segmented as background and the percentage of background segmented as oil spills, separately. 

\begin{table}[h]
 	\renewcommand\arraystretch{2.3}
 	\begin{center}
 	\tabcolsep 0.26in
 	\caption{CONFUSION MATRIX OF OIL SPILL SAR IMAGE SEGMENTATION.}
 	\begin{tabular}{c|cc}
		\hline
 		\hline
 		\multirow{2}{*}{}
		& oil & background\\
 		\hline
 		oil & 0.9879 & 0.0121 \\
 		\hline
 	    background& 0.0167  & 0.9833 \\
 		\hline
 		\hline
 	\end{tabular}
 	\label{confusion matrix for oil spill image segmentation}
 	\end{center}
 	\vspace*{-1.7mm}
 \end{table}

From the experimental validations in this subsection, we summarise that our proposed segmentation method outperforms both the circular initialisation and rectangle initialisation dependent methods. The effectiveness benefits from that our proposed segmentation network operates an automatic oil spill segmentation without manual initialisation. This provides an efficient and reliable segmentation for oil spill SAR images.

\subsection{Comparison with Logistic Regression Technique}
\begin{figure}[h]
 	\begin{center}
	\includegraphics[width=0.61\textwidth,height=0.47\textheight,center]{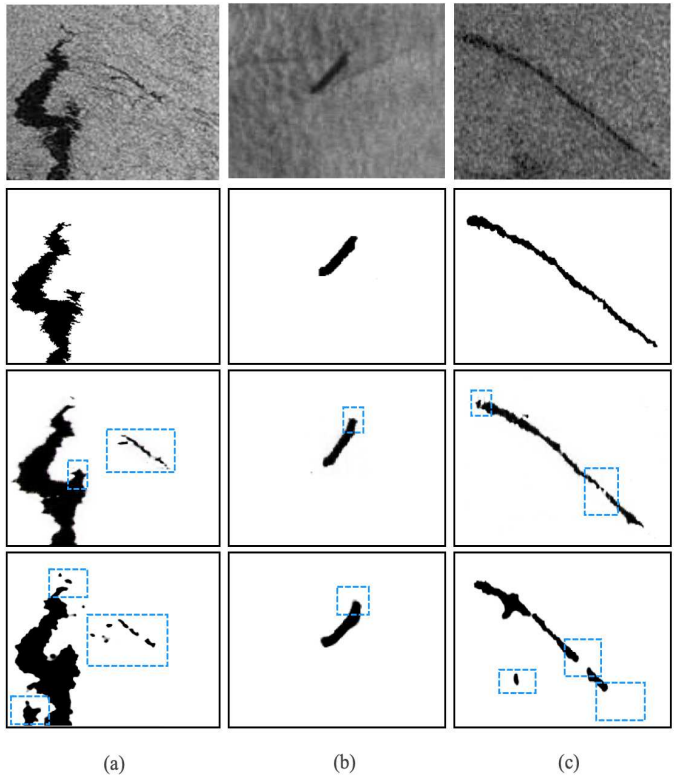}
	\end{center}
 	\vspace*{-4.7mm}
 	\caption{The segmentation of oil spill images using the proposed segmentation network and the logistic regression technique. Specifically, the top row from left to right includes the ERS-1 SAR, ERS-2 SAR and the Envisat ASAR oil spill images, respectively. The columns (a) to (c) from top to bottom are the original oil spill images, the ground-truth segmentation, the segmentation of the proposed segmentation network and the logistic regression technique, respectively. The blue dashed line boxes are utilised to indicate the incorrect segmentations.}
   \label{comparison with logistic regression model}
 \end{figure}

In image processing, the logistic regression strategy is often used for classification problems, and in the implementation of image segmentation, the logistic regression model is also popularly used \cite{li2011spectral}. Thus, to demonstrate the effectiveness of our proposed segmentation network for oil spill image segmentation, we here conduct further experimental evaluation by comparing its segmentation with that of the logistic regression technique \cite{khurshid2014segmentation}. Specifically, to compare the segmentation performance of our proposed segmentation network and the employed logistic regression methodology, we exploit them to segment ERS-1 SAR, ERS-2 SAR and Envisat ASAR oil spill images simultaneously, and the segmentation results are shown in Fig. \ref{comparison with logistic regression model}. In this figure, the top row from left to right includes the ERS-1 SAR, ERS-2 SAR and Envisat ASAR oil spill images, respectively, and columns (a)-(c) from top to bottom are the original oil spill images, the ground truth segmentation, the segmentation with our proposed segmentation network and the logistic regression model, respectively. The blue dashed line boxes are utilised to indicate the incorrect segmentations. Comparing the segmentation results of the employed segmentation methods with the ground truth segmentation, it is clear that our proposed segmentation network achieves a more accurate oil spill image segmentation result.

\subsection{Segmentation of Oil Spill SAR Images include Look-alikes}
\begin{figure}[h]
 	\begin{center}
	\includegraphics[width=0.51\textwidth,height=0.47\textheight,center]{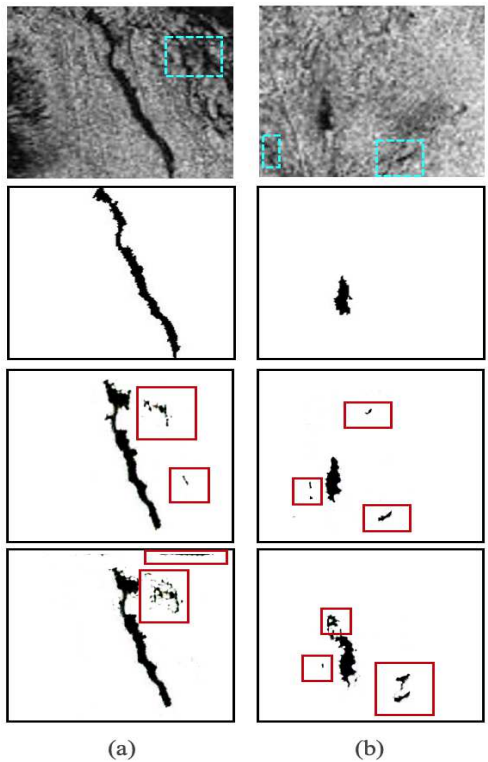}
	\end{center}
 	\vspace*{-8.6mm}
 	\caption{The segmentation of oil spills in ERS-1 SAR images. Specifically,  in columns (a) to (b) from top to bottom show the original SAR images which include look-alikes, the ground-truth segmentation, the segmentation of our proposed method and the GAN, respectively. The green boxes are exploited to indicate the look-alikes in SAR images, and the red boxes are utilised to indicate the incorrect segmentation of look-alikes in the segmentation outputs.}
   \label{segmentation ERS-1 SAR images include look-alikes}
 \end{figure}

\begin{figure}[h]
 	\begin{center}
	\includegraphics[width=0.51\textwidth,height=0.47\textheight,center]{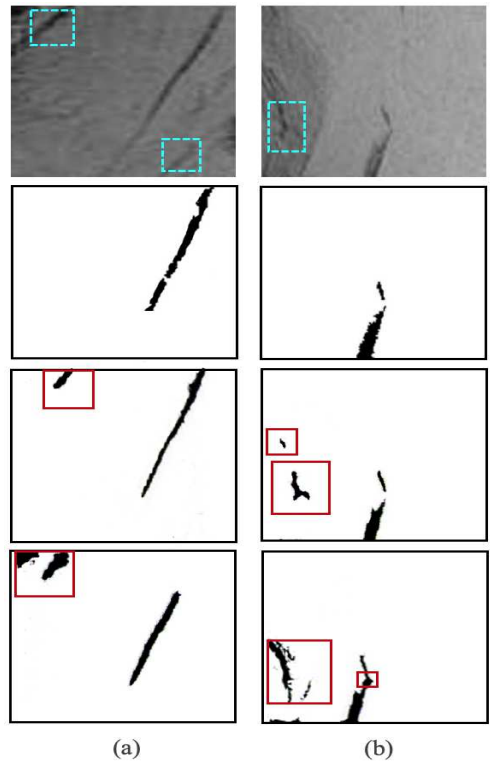}
	\end{center}
 	\vspace*{-8.6mm}
 	\caption{The segmentation of oil spills in ERS-2 SAR images. Specifically, in columns (a) to (b) from top to bottom show the original SAR images which include look-alikes, the ground-truth segmentation, the segmentation of our proposed method and the GAN, respectively. The green boxes are exploited to indicate the look-alikes in SAR images, and the red boxes are utilised to indicate the incorrect segmentation of look-alikes in the segmentation outputs.}
   \label{segmentation ERS-2 SAR images include look-alikes}
 \end{figure}

In the operation of oil spill SAR image segmentation, we should emphasize that the oil spill SAR images include look-alikes, and this makes difficulty for performing accurate oil spill segmentation. To address this problem, we here conduct further evaluation on our proposed segmentation network to examine its performance in segmenting SAR images including look-alikes. Specifically, in the experimental work, we utilise ERS-1 and ERS-2 SAR images as the experimental dataset, and the segmentation results for ERS-1 and ERS-2 SAR image segmentation are shown in Figs. \ref{segmentation ERS-1 SAR images include look-alikes} and \ref{segmentation ERS-2 SAR images include look-alikes}, respectively. The segmentation results show that our proposed method outperforms GAN for the segmentation of SAR images including look-alikes. 

In details, to examine the interference of look-alikes to the segmentation of oil spills, the ERS-1 and ERS-2 SAR images we utilised in the experimental work include different areas of look-alikes in the images. Particularly, in Fig. \ref{segmentation ERS-1 SAR images include look-alikes}, it shows that the original SAR images are of heavy look-alikes in large areas, while Fig. \ref{segmentation ERS-2 SAR images include look-alikes} shows that the original SAR images include a relative small area of look-alikes. Examining the segmentation results in these two figures, together with the segmented oil spill areas, the look-alikes are also partially segmented, and this degrades the segmentation performance of the employed segmentation methods. Thus, from the experimental evaluation on ERS-1 and ERS-2 SAR images including look-alikes, we can obtain that the look-alikes included in SAR images make it difficult to achieve effective oil spill segmentation. Besides, comparing the segmentation results from our proposed method and GAN with the ground truth segmentations, it is clear that our proposed method achieves higher accuracy in oil spill image segmentation. The optimal systematic performance mainly benefits from two factors. The first one is the training dataset. The training dataset provides supervisory information for the proposed system to push the segmentation towards the oil spill areas. The other one (also is the most important one) is the construction of the segmentation network. To conduct effective oil spill image segmentation, we construct our proposed oil spill image segmentation network by considering the intrinsic distribution of backscatter values in SAR images and incorporate the intrinsic distribution into our segmentation model. The intrinsic distribution originates from SAR imagery, describing the physical characteristics of oil spills. Thus, in the training process, the incorporated intrinsic distribution guides efficient learning for optimal latent feature inference, and benefiting from this, the proposed segmentation network achieves effective oil spill image segmentation. 

\subsection{Comprehensive Evaluations for Oil Spill SAR Image Segmentation}
\begin{figure*}[htbp]
	\begin{center}
	\includegraphics[width=1.23\textwidth,height=0.94\textheight,center]{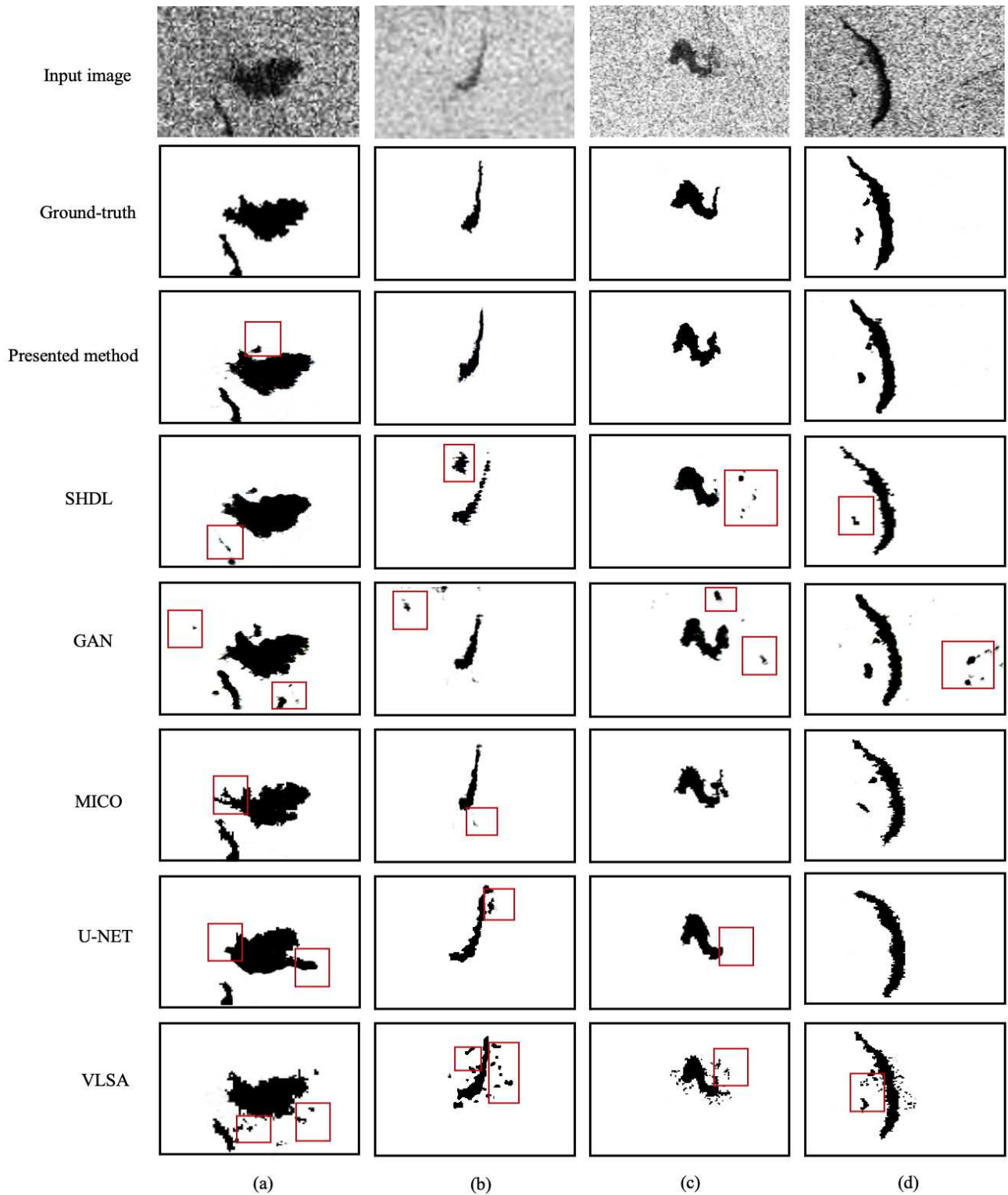}
	\end{center}
	\vspace*{-16.0mm}
	\caption{ERS-1 oil spill SAR image segmentation with the exploitation of different segmentation methodologies. Specifically, columns (a)-(d) show oil spill SAR images and their corresponding segmentation results with the implemented segmentation techniques. Particularly, from top to bottom in each column are the original oil spill SAR image, the ground-truth segmentation, the segmentation with our proposed method, SHDL, GAN, MICO, U-NET and the VLSA segmentation methodologies respectively. The red boxes are utilised to indicate the incorrect segmentations.}
  \label{ERS-1 oil spill SAR image segmentation}
\end{figure*}

We evaluate the segmentation performance of our proposed method by comparing its segmentation with several representative deep neural networks, region- and edge-based segmentation strategies, and the initialisation dependent methods in Sections \ref{initial segmentation results comparison} and \ref{comparison with the initialisation dependent method}, respectively, and the segmentation results validate the effectiveness of our proposed method. Particularly, the segmentation evaluations operated in Sections \ref{initial segmentation results comparison} and \ref{comparison with the initialisation dependent method} are conducted over different types of oil spill SAR images simultaneously. Thus, to evaluate the segmentation of our proposed method more comprehensively, we here conduct the segmentation validation from different metrics. Particularly, in this subsection, we exploit oil spill SAR images from different specific satellites, which include ERS-1 and ERS-2 satellites and the Envisat satellite. Moreover, to address the segmentation for irregular oil spill areas, we exploit the SAR images that contain oil spill areas with irregular shapes, and this enables a detailed examination for the segmentation capability of our proposed segmentation method. Specifically, in the experimental work, the comparison segmentation techniques employed include the squared helinger divergence learning (SHDL) method \cite{yu2018oil}, GAN, the multiplicative intrinsic component optimisation (MICO) technique \cite{li2014multiplicative}, U-Net neural network \cite{ronneberger2015u} and the variational level set approach (VLSA) \cite{zhang2014variational}, and the segmentation results for ERS-1 oil spill SAR images are shown in Fig. \ref{ERS-1 oil spill SAR image segmentation}. The segmentation results for ERS-2 oil spill SAR and Envisat oil spill ASAR images are shown in the Supplementary. Examining these segmentation results, it is obvious that our proposed method achieves more effective oil spill SAR image segmentation.

\begin{figure}[h]
	\begin{center}
	\includegraphics[width=0.47\textwidth,height=0.23\textheight,center]{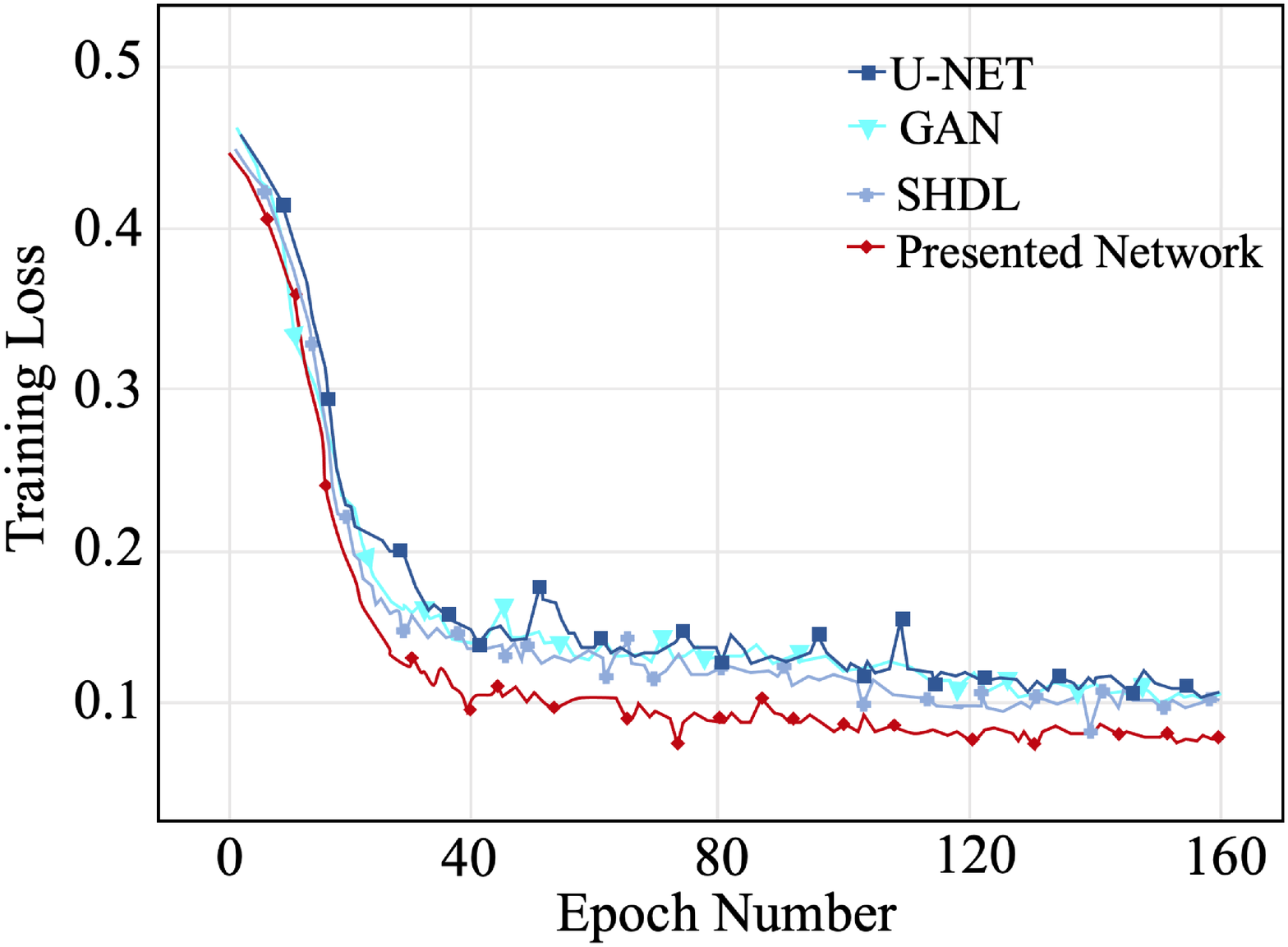}
	\end{center}
    \vspace*{-4.7mm}
	\caption{The learning curve of our proposed DGNet and the other employed segmentation neural networks.}
  \label{learning curve of the deep neural techniques}
\end{figure}

Moreover, to describe the efficiency of training for oil spill SAR image segmentation, we illustrate the training curve of our proposed segmentation network and the other implemented state-of-the-art deep neural networks in Fig. \ref{learning curve of the deep neural techniques}. As shown in this figure that our proposed method converges to a lower training loss with less epochs, compared against the other networks, and with the training epoch increasing, our proposed segmentation network fine tunes the segmentation map and thus operates accurate segmentation of oil spills. 

Additionally, to further evaluate the segmentation performance of our proposed method, together with the qualitative segmentation validation, we here quantitatively evaluate the segmentation in terms of the segmentation accuracy, and the segmentation accuracy of the implemented segmentation techniques (including our proposed DGNet) for ERS-1 oil spill SAR image segmentation is shown in Table \ref{ERS-1 oil spill SAR imagess segmentation accuracy}. The Tables in the Supplementary show the segmentation accuracy of ERS-2 and Envisat oil spill SAR images. From these Tables, it is clear that our proposed DGNet achieves higher accuracy for oil spill SAR image segmentation.

\begin{table*}[htbp]
	\renewcommand\arraystretch{1.8}
	\begin{center}
	\tabcolsep 0.17in
	\caption{SEGMENTATION ACCURACY OF ERS-1 OIL SPILL SAR IMAGES.}
	\begin{tabular}{c|cccccc}
		\hline
		\hline
		\multirow{2}{*}{}{\diagbox[innerwidth=2.6cm]{Image}{Method}}
		&VLSA & U-NET & MICO & GAN &SHDL & Our Method\\
		\hline
		\hline
		(a) & 0.9258 & 0.9103 & 0.9835 & 0.9749 & 0.9821 & \textbf{0.9847} \\
		\hline
		(b) & 0.8903 & 0.9087 & 0.9635 & 0.8967 & 0.8875 & \textbf{0.9886} \\
		\hline
		(c) & 0.9033 & 0.9096 & 0.9658 & 0.9107 & 0.8903 & \textbf{0.9729} \\
		\hline
		(d) & 0.9158 & 0.9476 & 0.9679 & 0.9055 & 0.9806 &  \textbf{0.9887} \\
		\hline
		\hline
	\end{tabular}
	\label{ERS-1 oil spill SAR imagess segmentation accuracy}
	\end{center}
\end{table*}

In the implementation of oil spill SAR image segmentation, an accurate segmentation is that the segmentation maps tightly follow the ground-truth segmentations. Thus, to validate the segmentation performance of our proposed method one step further, we examine the closeness between the produced segmentation and the ground-truth segmentation with respect to the region fitting rate (RFR), which is given below:
\begin{equation*}\label{region fitting rate computation}
\text{RFR} = \frac{ |G_{R}\bigcap S_{R}|card}{|G_{R}\bigcup S_{R}|card}
\end{equation*}
where $G_{R}$ and $S_{R}$ represent the regions of ground truth segmentation and the produced segmentation, respectively, and $|\cdot|_{card}$ indicates the cardinality of sets. The computation shows that a higher RFR represents a more accurate oil spill SAR image segmentation, and when the produced segmentation closely matches the ground truth segmentation, the RFR approaches one. To describe the overall performance of the segmentation techniques, in the experimental work, we exploit over 160 oil spill SAR images as the test dataset, and the RFR that correlates each segmentation methodology is shown in Fig. \ref{region fitting rate computation}. As shown in this figure that our proposed segmentation method achieves a higher level of RFR. This demonstrates that our proposed method performs the segmentation that closely follows the ground-truth segmentation, compared against the other implemented segmentation techniques.

\begin{figure*}[htbp]
  \begin{center}
  %
  \subfigure[]{
  \includegraphics[width=0.31\textwidth,height=0.17\textheight]{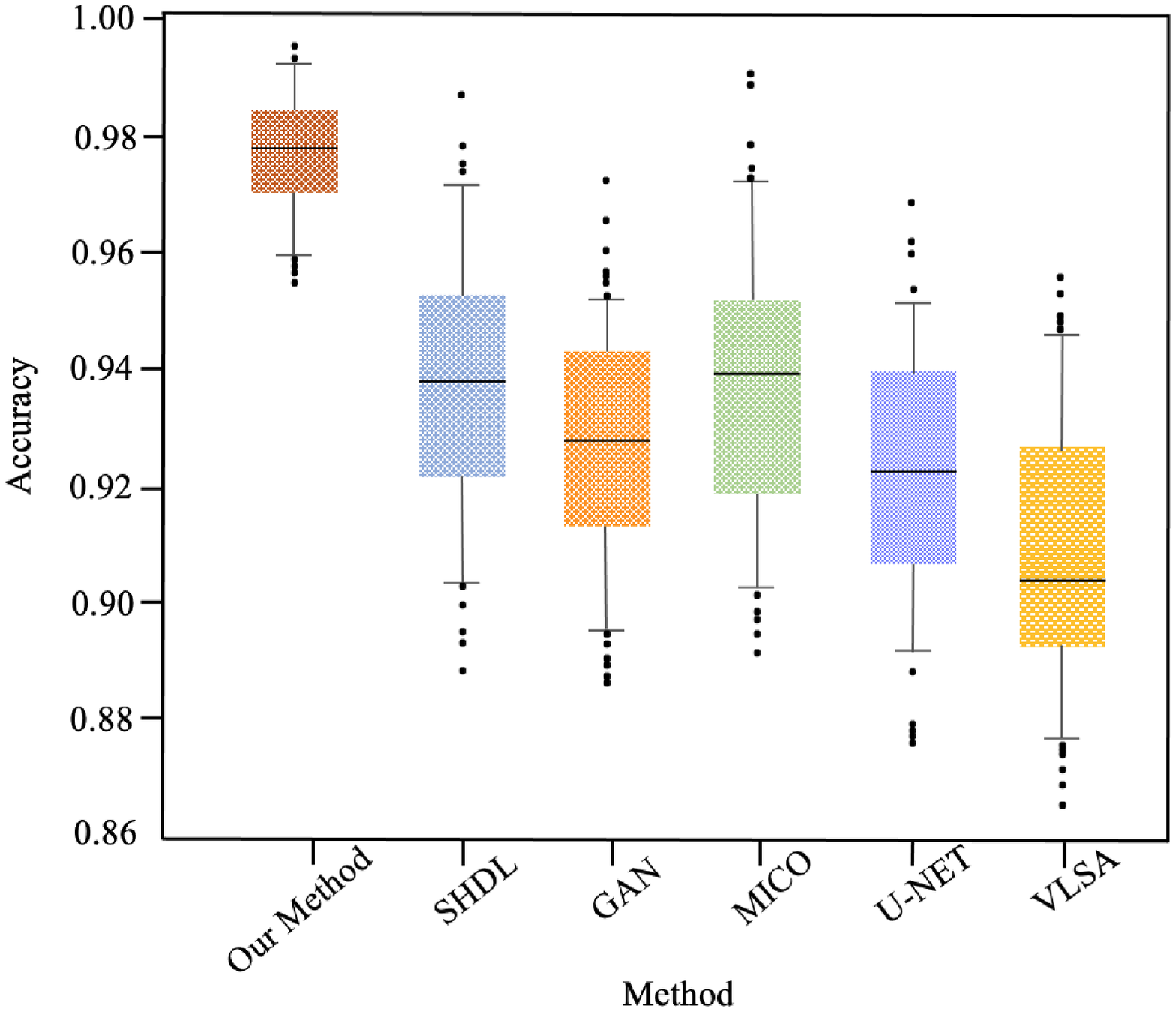}}
  \subfigure[]{
  \includegraphics[width=0.31\textwidth,height=0.17\textheight]{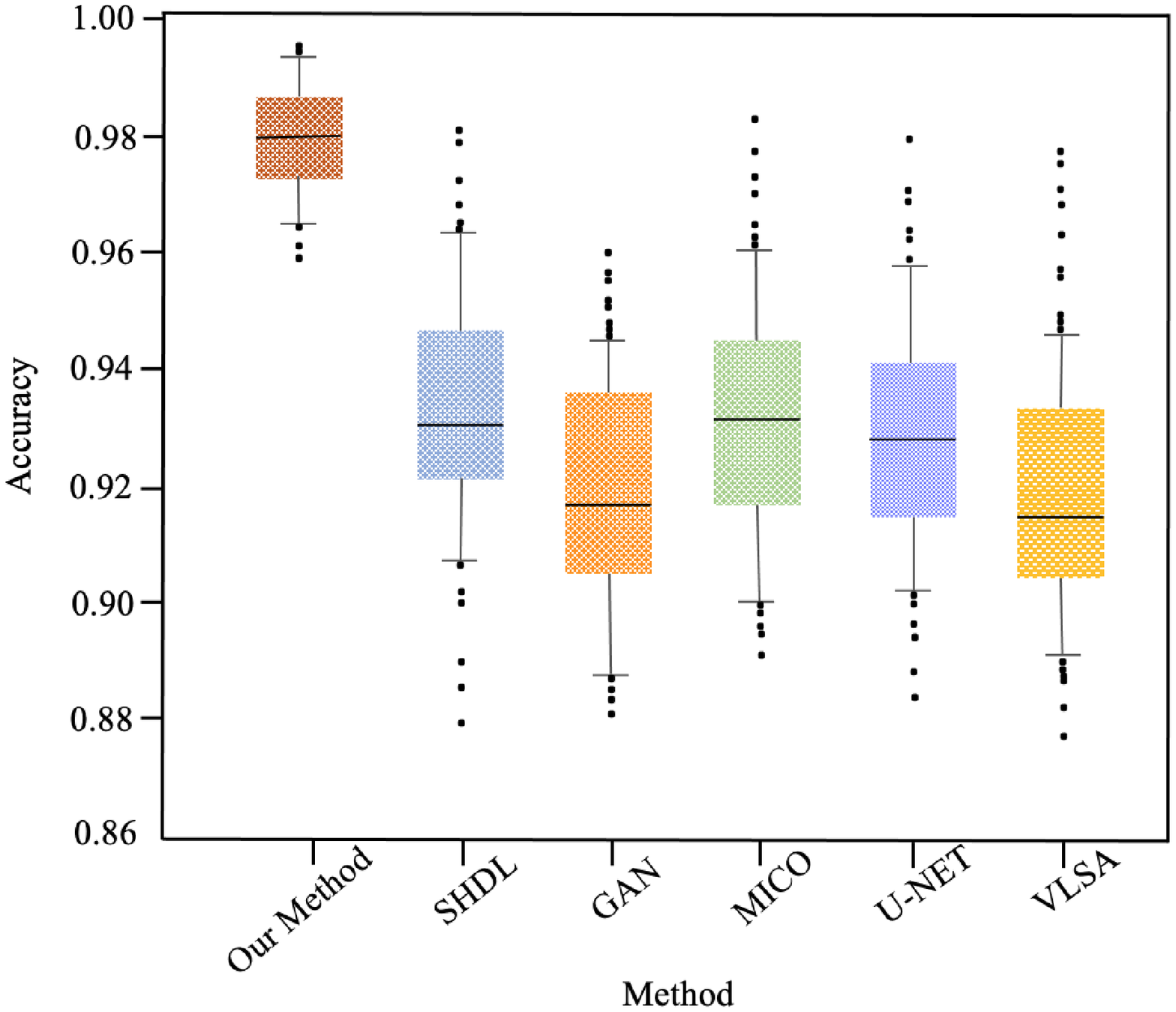}}
  \subfigure[]{
  \includegraphics[width=0.31\textwidth,height=0.17\textheight]{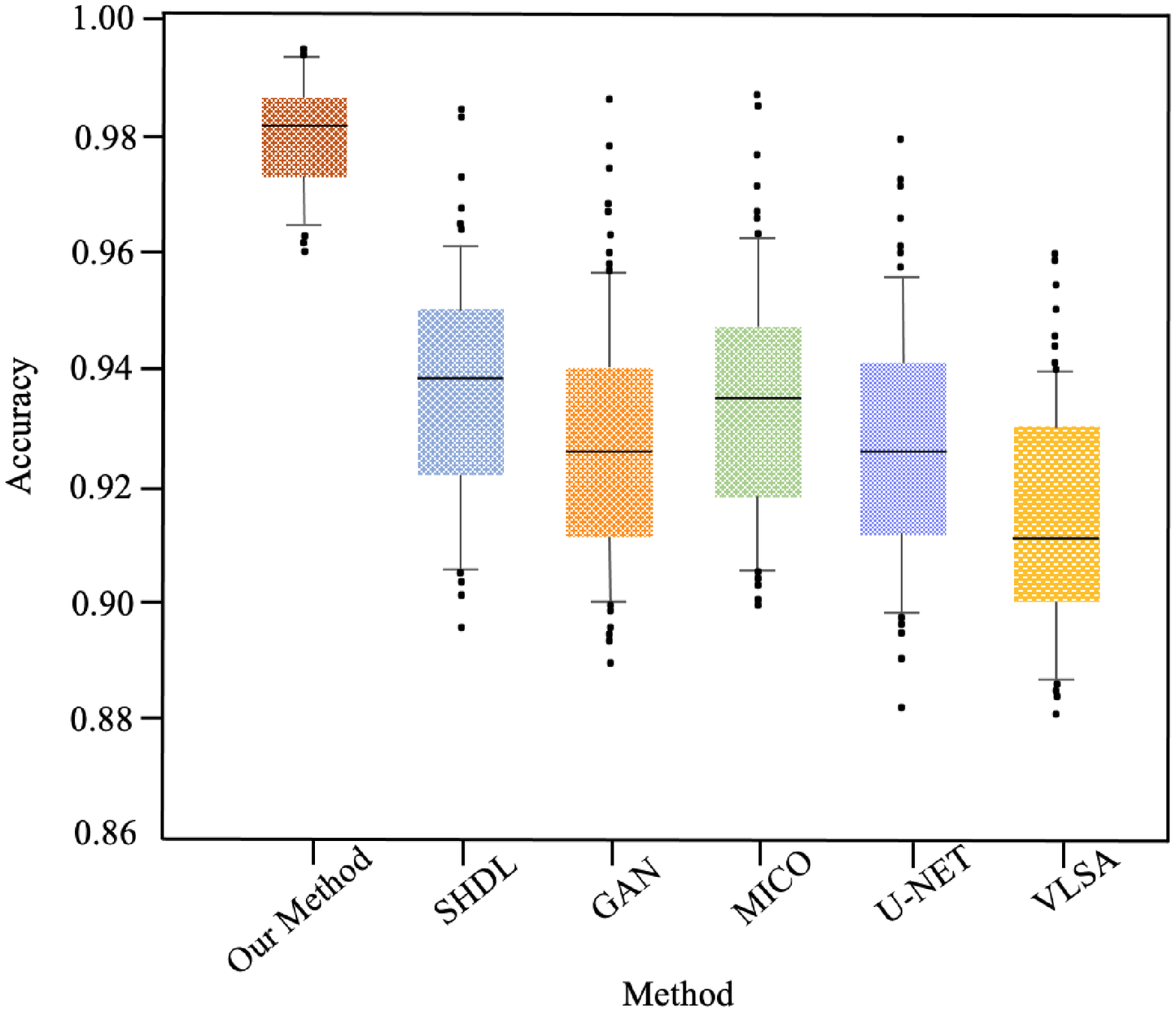}}
  \end{center}
  \vspace*{-2mm}
  \caption{Segmentation accuracy distributions with respect to different types of oil spill images. Specifically, (a), (b) and (c) illustrate the distributions of the segmentation accuracy that corresponds to each method for ERS-1, ERS-2 and Envisat oil spill image segmentation respectively. The rectangles corresponding to the strategies depict the accuracy values distributed between the minimum and maximum values (these two values are shown with bars on bottom-up sides of the rectangles), and the black dots are the segmentation outliers that fall out the interval of the two bars.}
  \label{accuracy with standard deviation} 
\end{figure*}

In addition, in the implementation of oil spill SAR image segmentation, to further validate the performance of our proposed method, we wish to attain statistical results as a quantitative way for describing the segmentation accuracy and stability of the methods, and the statistical results are illustrated in Fig. \ref{accuracy with standard deviation}. Specifically, Fig. \ref{accuracy with standard deviation}(a), (b) and (c) show the segmentation distribution ranges of the methods for ERS-1 SAR, ERS-2 SAR and Envisat ASAR oil spill image segmentation, respectively. The coloured boxes in each sub-figure are employed to depict the clustered accuracy values, where the bars on the bottom-up sides of the boxes are the levels of the minimum and maximum values, and the black dots are the segmentation outliers that fall out the interval of the two bars. This figure shows that our proposed method achieves higher segmentation accuracy with a smaller number of outliers, compared against the other segmentation methodologies. This demonstrates that our proposed method achieves more accurate and stable oil spill SAR image segmentation. 



\begin{figure*}[htbp]
  \begin{center}
  %
  \subfigure[]{
  \includegraphics[width=0.31\textwidth,height=0.16\textheight]{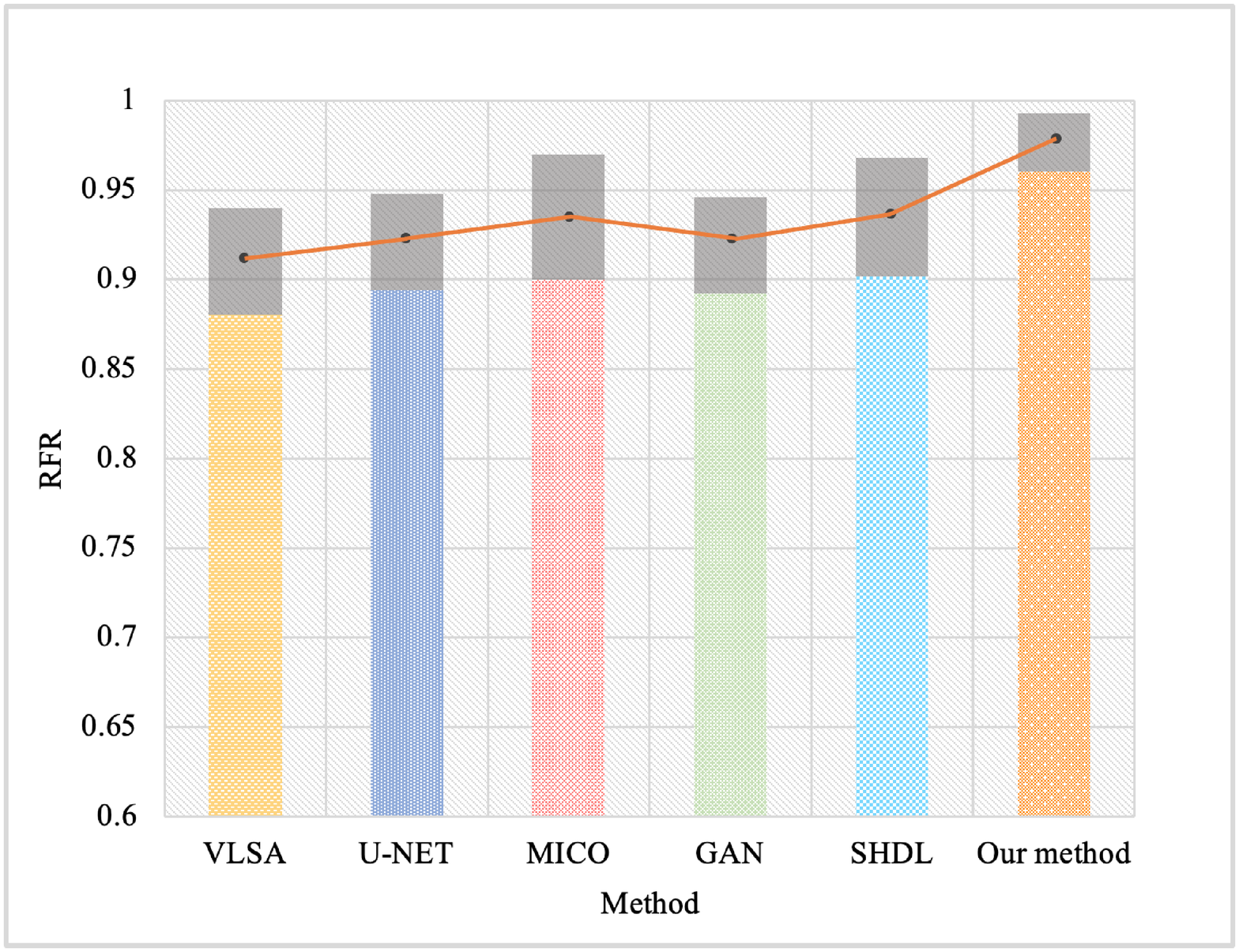}}
  \subfigure[]{
  \includegraphics[width=0.31\textwidth,height=0.16\textheight]{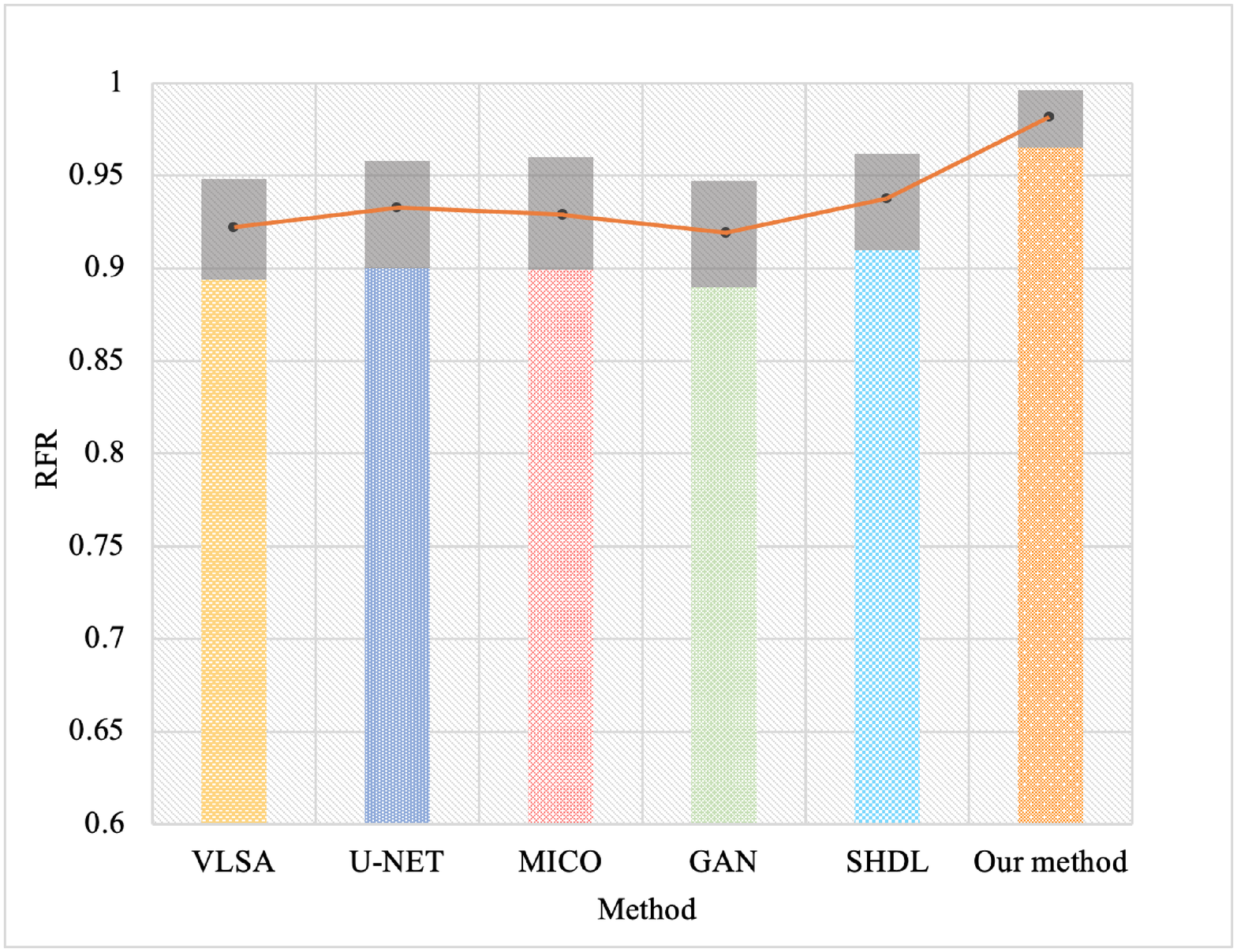}}
  \subfigure[]{
  \includegraphics[width=0.31\textwidth,height=0.16\textheight]{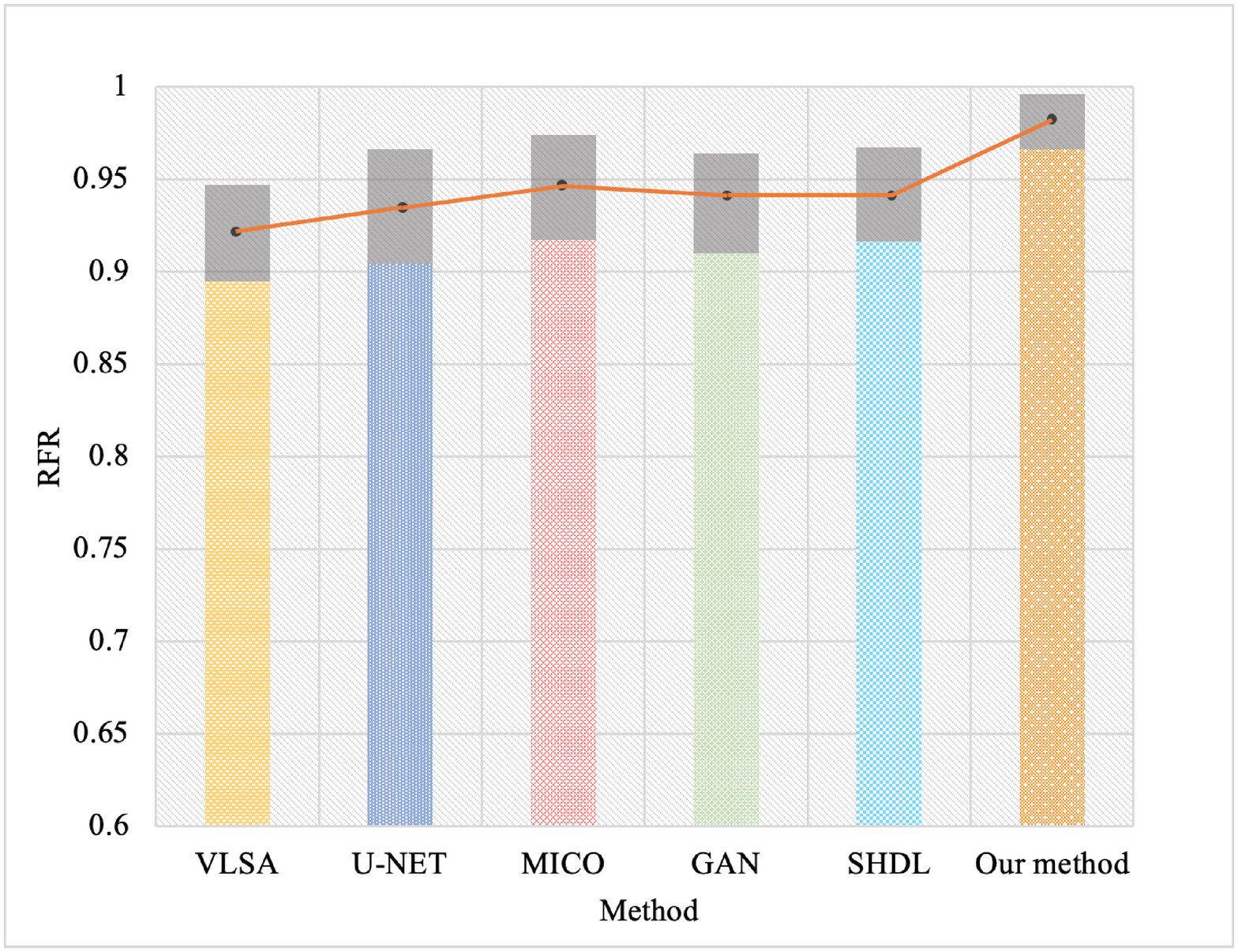}}
  \end{center}
\vspace*{-2mm}
  \caption{The segmentation performance with respect to region fitting rate (RFR) of oil spill SAR image segmentation with the exploitation of different segmentation methods. In this figure, from (a) to (c) indicate the RFR of the employed methods for ERS-1, ERS-2 oil spill SAR and the Envisat oil spill ASAR image segmentation respectively, and the columns in each sub-figure indicate the RFR of the methods, with the top grey sections utilised to mark the standard deviations, and the line that connects the mean points in the grey sectors shows the trend of the segmentation performance of the alternative methods.}
  \label{region fitting rate computation} 
\end{figure*}

\section{Conclusion}
We have presented our novel marine oil spill SAR image segmentation framework named DGNet, which presents effective oil spill segmentation with the incorporation of the intrinsic distribution of backscatter values in oil spill SAR images. Specifically, our proposed segmentation framework was structured with an inference module and a generative module that work cooperatively in an interactive manner, and in the training for oil spill SAR image segmentation, the inference module learns the representation of latent feature variable inference for SAR images, and the generative module generates oil spill segmentation maps by drawing the latent feature variables from the inference outputs. Thus, to render optimal latent feature variable inference for generating accurate oil spill segmentation maps, we investigated the intrinsic distribution of backscatter values in oil spill SAR images and incorporate it into our segmentation model. The intrinsic distribution originates from SAR imagery for maritime scenes, and thus it provides physical characteristics of oil spills. Thus, in the training process, the incorporated intrinsic distribution guides efficient learning of optimal latent feature variable inference for effective oil spill segmentation. The efficient learning enables the training of our proposed DGNet with a small number of image data. This is economically beneficial to oil spill segmentation where the availability of oil spill SAR image data is limited in practice. Moreover, benefiting from optimal latent feature variable inference, the generative module generates accurate oil spill segmentation including the segmentation for irregular oil spill areas. Extensive experimental evaluations from different metrics have validated the effectiveness of our proposed DGNet for different types of oil spill SAR image segmentation, compared against several state-of-the-art segmentation methodologies. 


%





\ifCLASSOPTIONcaptionsoff
  \newpage
\fi

%






\bibliographystyle{IEEEtran}
\bibliography{reference}

\end{document}


\markboth{IEEE TRANSACTIONS ON GEOSCIENCE AND REMOTE SENSING}%
{Shell \MakeLowercase{\textit{et al.}}: Bare Demo of IEEEtran.cls for IEEE Journals}


\section{Supplementary}
 


\begin{figure}[H]
	\begin{center}
	\includegraphics[width=1.23\textwidth,height=0.94\textheight,center]{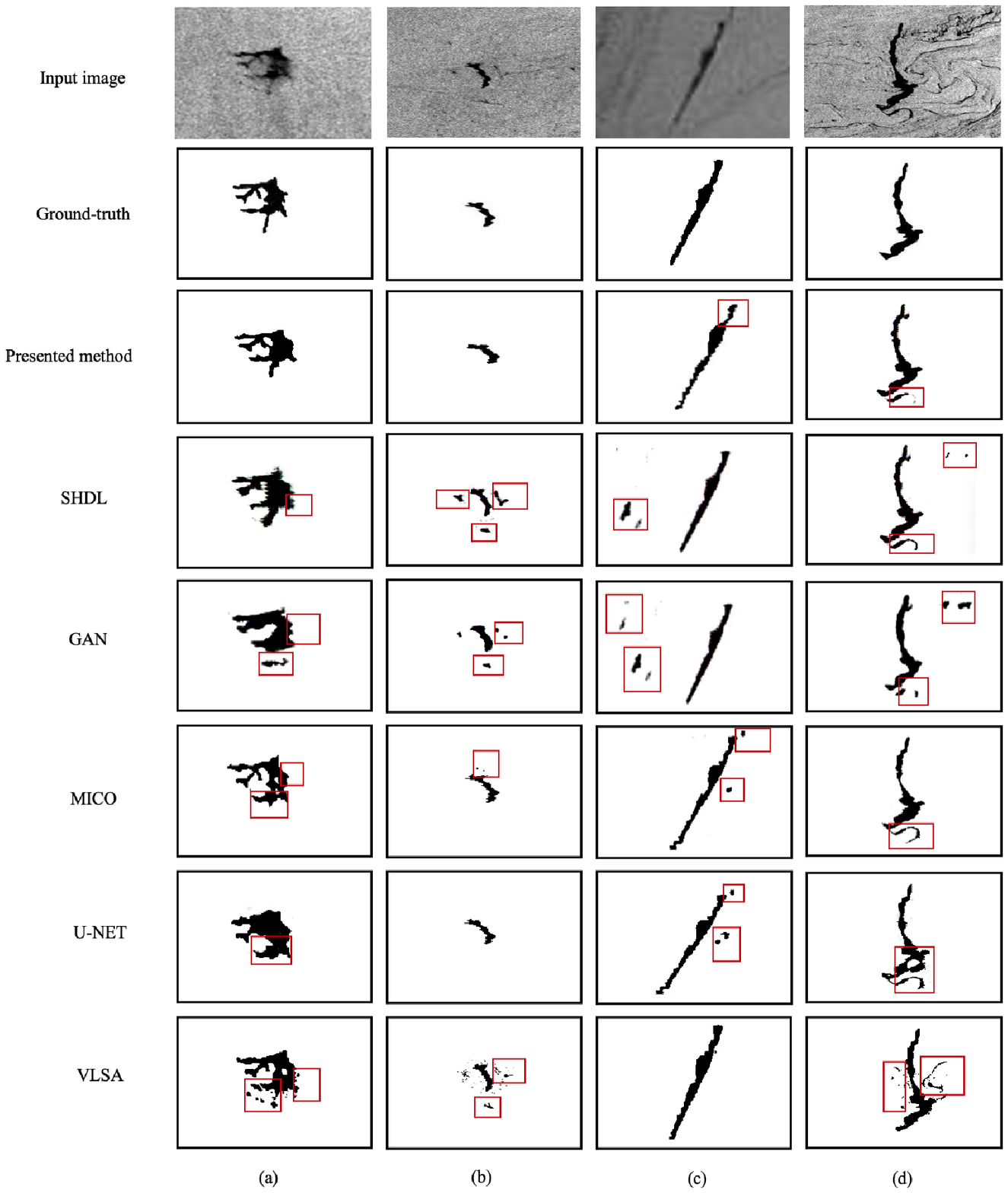}
	\end{center}
	\vspace*{-16.0mm}
	\caption{ERS-2 oil spill SAR image segmentation with the exploitation of different segmentation methodologies. Specifically, columns (a)-(d) show oil spill SAR images and their corresponding segmentation results with the implemented segmentation techniques. Particularly, from top to bottom in each column are the original oil spill SAR image, the ground-truth segmentation, the segmentation with our proposed method, SHDL, GAN, MICO, U-NET and the VLSA segmentation methodologies respectively. The red boxes are utilised to indicate the incorrect segmentations.}
  \label{ERS-2 oil spill SAR image segmentation}
\end{figure}

\begin{figure}[H]
	\begin{center}
	\includegraphics[width=1.22\textwidth,height=0.94\textheight,center]{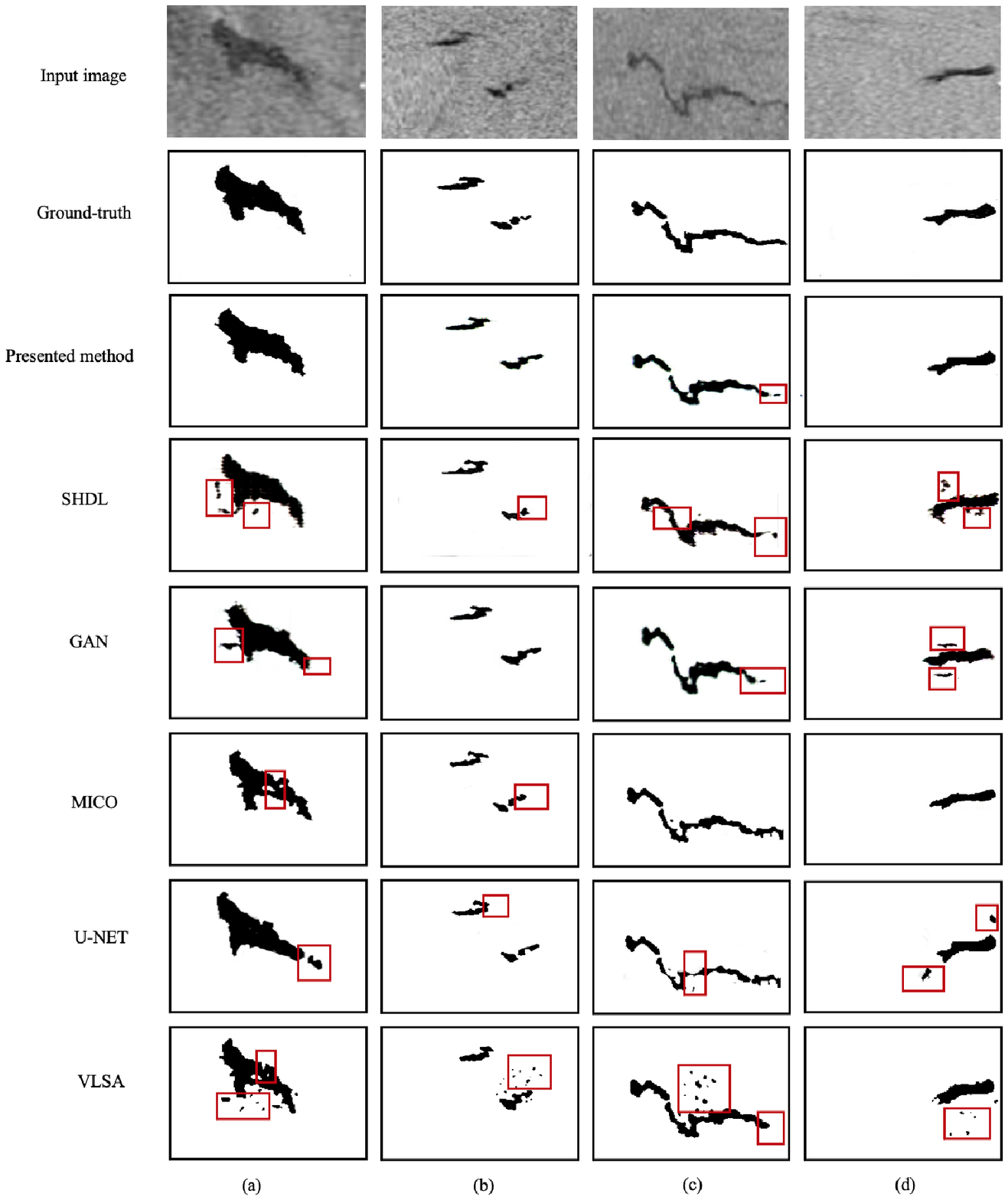}
	\end{center}
	\vspace*{-16.0mm}
	\caption{Envisat oil spill ASAR image segmentation with the exploitation of different segmentation methodologies. Specifically, columns (a)-(d) show oil spill ASAR images and their corresponding segmentation results with the implemented segmentation techniques. Particularly, from top to bottom in each column are the original oil spill ASAR image, the ground-truth segmentation, the segmentation with our proposed method, SHDL, GAN, MICO, U-NET and the VLSA segmentation methodologies respectively. The red boxes are utilised to indicate the incorrect segmentations}
  \label{Envisat oil spill ASAR image segmentation}
\end{figure}

 \begin{table}[H]
 	\renewcommand\arraystretch{1.8}
 	\begin{center}
 	\tabcolsep 0.17in
 	\caption{SEGMENTATION ACCURACY OF ERS-2 OIL SPILL SAR IMAGES.}
 	\begin{tabular}{c|cccccc}
 		\hline
 		\hline
 		\multirow{2}{*}{}{\diagbox[innerwidth=2.6cm]{Image}{Method}}
 		&VLSA & U-NET & MICO & GAN &SHDL & Our Method\\
 		\hline
 		\hline
 		(a) & 0.9179 & 0.8933 & 0.9586 & 0.9058 & 0.96895 & \textbf{0.9953} \\
 		\hline
 		(b) & 0.9675 & 0.9863 & 0.9659 & 0.9187 & 0.8952 & \textbf{0.9947} \\
 		\hline
 		(c) & 0.9752 & 0.9606 & 0.9621 & 0.9096 & 0.9408 & \textbf{0.9866} \\
 		\hline
 		(d) & 0.9423 & 0.9308 & 0.9649 & 0.9453 & 0.9658 & \textbf{0.9743}  \\
 		\hline
 		\hline
 	\end{tabular}
	\label{ERS-2 oil spill SAR imagess segmentation accuracy}
  \end{center}
 \end{table}

 \begin{table}[H]
 	\renewcommand\arraystretch{1.8}
 	\begin{center}
 	\tabcolsep 0.17in
 	\caption{SEGMENTATION ACCURACY OF ENVISAT OIL SPILL ASAR IMAGES.}
 	\begin{tabular}{c|cccccc}
 		\hline
 		\hline
 		\multirow{2}{*}{}{\diagbox[innerwidth=2.6cm]{Method}{Image}}
 		&VLSA & U-NET & MICO & GAN &SHDL & Our Method\\
 		\hline
		\hline
 		(a) & 0.9179 & 0.9436 & 0.9575 & 0.9568 & 0.9643 & \textbf{0.9887}  \\
 		\hline
		(b) & 0.9265 & 0.9859 & 0.9876 & 0.9894 & 0.9885 & \textbf{0.9906} \\
 		\hline
 		(c) & 0.9283 & 0.9659 & 0.9758 & 0.9429 & 0.9769 & \textbf{0.9875} \\
 		\hline
 		(d) & 0.9335 & 0.9568 & 0.9939 & 0.9358 & 0.9475 & \textbf{0.9947} \\
 		\hline
 		\hline
 	\end{tabular}
 	\label{Envisat oil spill ASAR images segmentation accuracy}
 	\end{center}
 \end{table}


%





\ifCLASSOPTIONcaptionsoff
  \newpage
\fi

%






\bibliographystyle{IEEEtran}